\documentclass[10pt,journal]{IEEEtran}
\usepackage{diagbox}
\usepackage{amssymb}
\usepackage{color, colortbl}
\usepackage{color}
\usepackage[T1]{fontenc}
\usepackage{textcomp}
\usepackage{textgreek}
\usepackage{caption}
\usepackage[table]{xcolor}
\usepackage{bm}
\usepackage[colorlinks]{hyperref}
\usepackage{enumitem}
\usepackage{sidecap}

\newcommand{\A}[1]{{\textcolor{black}{#1}}}
\newcommand{\response}[1]{{\textcolor{black}{#1}}}
\usepackage{algcompatible}

\sidecaptionvpos{figure}{t}

\usepackage{cite}
\usepackage[linesnumbered,ruled,vlined]{algorithm2e}
\usepackage{multicol}
\usepackage{multicol}
\usepackage{multirow}
\usepackage[export]{adjustbox}
\usepackage{color}
\usepackage{float}
\usepackage[caption = false]{subfig}
\usepackage{pifont}

\newcommand{\cmark}{\ding{51}}%

\usepackage{amsmath}
\ifCLASSINFOpdf
\else
\fi

\DeclareMathOperator*{\argmax}{argmax}
\begin{document}

\title{Image Super-Resolution as a Defense Against Adversarial Attacks}

\author{Aamir Mustafa, Salman H. Khan, Munawar Hayat, Jianbing Shen and Ling Shao
\thanks{All authors are with Inception Institute of Artificial Intelligence, UAE.}
\thanks{A.~Mustafa and M.~Hayat are also with University of Canberra, Australia.}
\thanks{S.~H.~Khan is also with Data61-CSIRO, Canberra, Australia.}
\thanks{J.~Shen is also with Beijing Institute of Technology, China.}
\thanks{Contact: firstname.lastname@inceptioniai.org}}

% \author{Michael~Shell,~\IEEEmembership{Member,~IEEE,}
%         John~Doe,~\IEEEmembership{Fellow,~OSA,}
%         and~Jane~Doe,~\IEEEmembership{Life~Fellow,~IEEE}% <-this % stops a space
% \thanks{M. Shell was with the Department
% of Electrical and Computer Engineering, Georgia Institute of Technology, Atlanta,
% GA, 30332 USA e-mail: (see http://www.michaelshell.org/contact.html).}% <-this % stops a space
% \thanks{J. Doe and J. Doe are with Anonymous University.}% <-this % stops a space
% \thanks{Manuscript received April 19, 2005; revised August 26, 2015.}}

% % The paper headers
% \markboth{Journal of \LaTeX\ Class Files,~Vol.~14, No.~8, August~2015}%
% {Shell \MakeLowercase{\textit{et al.}}: Bare Demo of IEEEtran.cls for IEEE Journals}

\maketitle

\begin{abstract}
Convolutional Neural Networks have achieved significant success across multiple computer vision tasks. However, they are vulnerable to carefully crafted, human-imperceptible adversarial noise patterns which constrain their deployment in critical security-sensitive systems. This paper proposes a computationally efficient image enhancement approach that provides a strong defense mechanism to effectively mitigate the effect of such adversarial perturbations. We show that deep image restoration networks learn mapping functions that can bring \textit{off-the-manifold} adversarial samples onto the natural image manifold, thus restoring classification towards correct classes. A distinguishing feature of our approach is that, in addition to providing robustness against attacks, it simultaneously enhances image quality and retains models performance on clean images. Furthermore, the proposed method does not modify the classifier or requires a separate mechanism to detect adversarial images. The effectiveness of the scheme has been demonstrated through extensive experiments, where it has proven a strong defense in \textit{gray-box} settings. The proposed scheme is simple and has the following advantages: (1) it does not require any model training or parameter optimization, (2) it complements other existing defense mechanisms, (3) it is agnostic to the attacked model and attack type and (4) it provides superior performance across all popular attack algorithms. Our codes are publicly available at \url{https://github.com/aamir-mustafa/super-resolution-adversarial-defense}.
\end{abstract}

% no keywords

\IEEEpeerreviewmaketitle

\section{Introduction}
\label{sec:intro}

% \MH{to-dos:
% \begin{enumerate}
    % \item some defense mechanisms deteriorate/change image quality; which could affect their results even on clean/non-attacked images. I think we should show a comparison of that on full ILSVRC val set. That is, take clean images, pass them through respective defense algo; get classification. We could them make a point that SR proves to be an effective defense, while simultaneously retaining or even improving performance on clean images.
%   \end{enumerate}
% }

Success of Convolutional Neural Networks (CNNs) over the past several years has lead to their extensive deployment in a wide range of computer vision tasks \cite{jb1, jb2}, including image classification \cite{he2016deep, krizhevsky2012imagenet, simonyan2014very}, object detection \cite{girshick2015fast, ren2015faster}, semantic segmentation\cite{long2015fully,chen2018deeplab} and visual question answering\cite{santoro2017simple}. Not only limited to that, CNNs now play a pivotal role in designing many critical real-world systems, including self-driving cars\cite{santana2016learning} and models for disease diagnosis\cite{lee2017deep}, which necessitates their robustness in such situations. Recent works \cite{szegedy2013intriguing,43405,kurakin2016}, however, have shown that CNNs can easily be fooled by distorting natural images with small, well-crafted, human-imperceptible additive perturbations. These distorted images, known as \emph{adversarial examples}, have further been shown to be transferable across different architectures, e.g an adversarial example generated for an Inception v-3 model is able to fool other CNN architectures \cite{szegedy2013intriguing, akhtar2018threat}.

\begin{figure}[t]
  \centering
  \includegraphics[trim={3.5cm 3.5cm 5.5cm 5cm}, clip, width=0.5\textwidth]{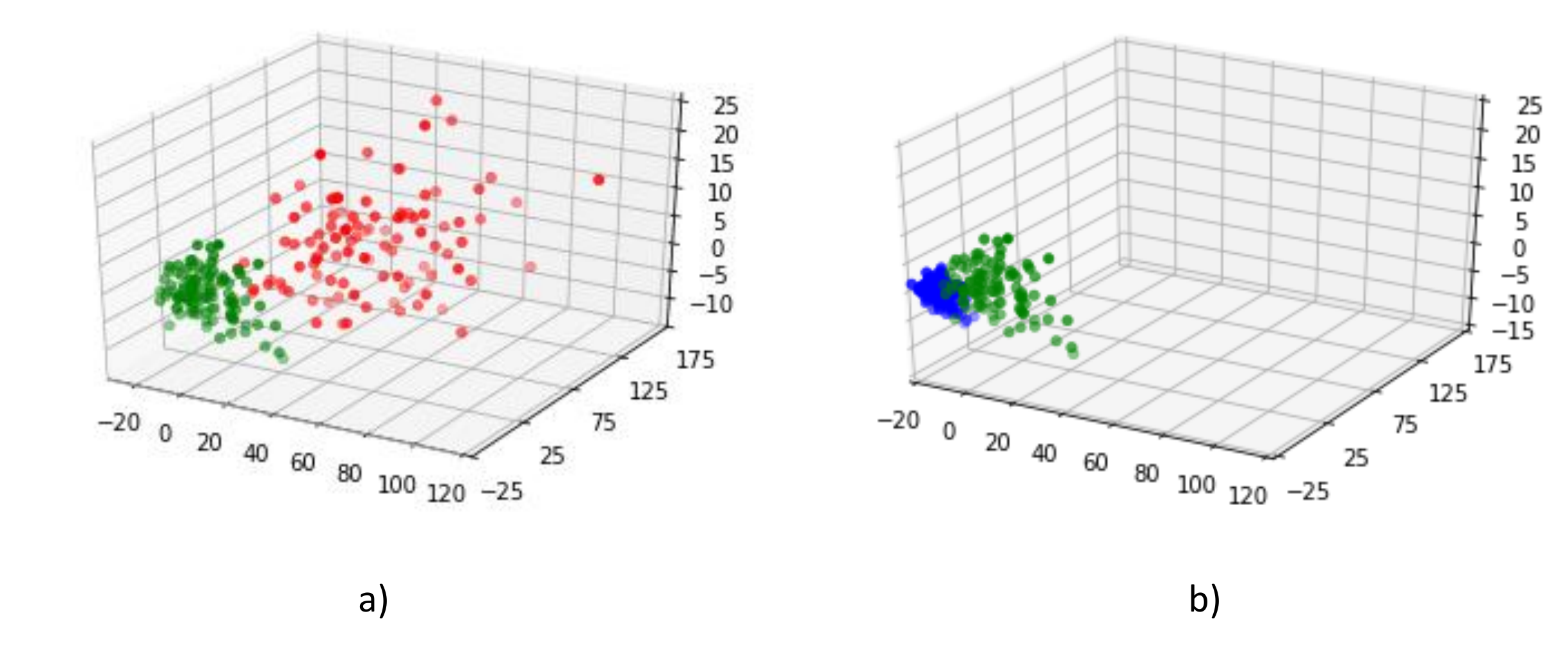}
  \caption{\footnotesize{a) A 3D plot showing  adversarial image features (red) and the corresponding clean image features (green). b) On the right, we show the features of the corresponding defended images (blue). The plot clearly shows that the super-resolution operation remaps the adversarial images to the natural image manifold, which otherwise lie off manifold. (100 randomly selected features projected to 3D space using principal component analysis are shown for better visualization)}} 
  \label{fig:manifold-assumption}
\end{figure}

Owing to the critical nature of security-sensitive CNN applications, significant research has been carried out to devise defense mechanisms against these vulnerabilities \cite{papernot2016distillation, xie2017mitigating, kurakin2016adversarial, tramer2017ensemble, mustafa2019adversarial, cao2017mitigating, metzen2017detecting, feinman2017detecting, meng2017magnet, prakash2018deflecting, liao2017defense, xie2018feature}. We can broadly categorize these defenses along two directions: the \textit{first} being model-specific mechanisms, which aim to regularize a specific model's parameters through adversarial training or parameter smoothing \cite{tramer2017ensemble, kurakin2016adversarial, papernot2016distillation, cisse2017parseval, liao2017defense}. Such methods often require differentiable transformations that are computationally demanding. Moreover these transformations are vulnerable to further attacks, as the adversaries can circumvent them by exploiting the differentiable modules. The \textit{second} category of defenses are model-agnostic. They mitigate the effect of adversarial perturbations in the input image domain by applying various transformations. Examples of such techniques include JPEG compression \cite{dziugaite2016study, das2017keeping}, foveation-based methods, which crop the image background \cite{luo2015foveation}, random pixel deflection \cite{prakash2018deflecting} and random image padding \& re-sizing  \cite{xie2017mitigating}. Compared with differentiable model-specific methods, most of the model-agnostic approaches are computationally faster and carry out transformations in the input domain, making them more favorable. However, most of these approaches lose critical image content when removing adversarial noise, which results in poor classification performance on non-attacked images. %This results in poor a major limitations of these approaches is the loss of information content in the image to overcome adversarial  most of the existing model-agnostic methods degrade image quality and result in information loss. % compared with model-specific, since they do not incur computationally expensive model training All of these methods however degrade the quality of image and may result in losing significant information.

This paper proposes a model-agnostic defense mechanism against a wide range of recently proposed adversarial attacks \cite{43405, kurakin2016, dong2018boosting, moosavi2016deepfool, carlini2017towards, xie2018improving} and does not suffer from information loss. Our proposed defense is based upon image super-resolution (SR), which selectively adds high frequency components to an image and removes noisy perturbations added by the adversary. We hypothesize that the learned SR models are generic enough to remap \textit{off-the-manifold} samples onto the natural image manifold (see Fig.~\ref{fig:manifold-assumption}). The effect of added noise is further suppressed by wavelet domain filtering and inherently minimized through a global pooling operation on the higher resolution version of the image. The proposed image super-resolution and wavelet filtering based defense results in a joint non-differentiable module, which can efficiently recover the original class labels for adversarially perturbed images.

The main contributions of our work are: 
\begin{enumerate}[leftmargin=0.25in]
\item Through extensive empirical evaluations, we show image super-resolution to be an effective defense strategy against a wide range of recently proposed state-of-the-art attacks in the literature \cite{43405, kurakin2016, dong2018boosting, moosavi2016deepfool, carlini2017towards, xie2018improving}. Using Class Activation Map visualizations, we demonstrate that super-resolution can successfully divert the attention of the classifier from random noisy patches to more distinctive regions of the attacked images (see Fig.~\ref{fig:fgsm} and ~\ref{fig:cw}).
\item Super-resolving an adversarial image projects it back to the natural image manifold learned by deep image classification networks.
\item Unlike existing image transformation based techniques, which introduce artifacts in the process of overcoming adversarial noise, the proposed scheme retains critical image content, and thus minimally impacts the classifier's performance on clean, non-attacked images.
    
% \item The proposed defense constitutes simple, computationally inexpensive transformations in the image domain and can easily complement other existing model-specific defense methods.

\item The proposed defense mechanism tackles adversarial attacks with no knowledge of the target model's architecture or parameters. This can easily complement other existing model-specific defense methods.
\end{enumerate}

Closely related to our approach are the Defense-GAN \cite{samangouei2018defense} and MagNet \cite{meng2017magnet}, which first estimate the manifold of clean data to detect adversarial examples and then apply a mapping function to reduce adversarial noise. Since they use generator blocks to re-create images, their studied case is restricted to small datasets (CIFAR-10, MNIST) with low-resolution images.  In contrast, our approach does not require any prior detection scheme and works for all types of natural images with a more generic mapping function.

Below, we first formally define the underlying problem (Sec.~\ref{sec:problem}), followed by a brief description of existing  adversarial attacks (Sec.~\ref{sec:adversrial}) and defenses (Sec.~\ref{sec:Adversarial Defenses}). We then present our proposed defense mechanism (Sec.~\ref{sec:purification}). The effectiveness of our proposed defense is then demonstrated through extensive experiments against state-of-the art adversarial attacks \cite{43405, kurakin2016, dong2018boosting, moosavi2016deepfool, carlini2017towards, xie2018improving} and comparison with other recently proposed model-agnostic defenses \cite{das2018shield,xie2017mitigating,guo2017countering,prakash2018deflecting} (see Section \ref{sec:experiments}). 

% In Section. \ref{sec:adversrial} we discuss about the established attacks techniques in literature. Section \ref{sec:purification} describes our defense mechanism and its components. In Section \ref{sec:related_work} we discuss closely related defense techniques with which we compare our defense scheme.  In Section \ref{sec:experiments} we lay out our experimental paradigm and results. Finally we conclude in Section \ref{sec:conclusion}.

\section{Background}
Here we introduce popular adversarial attacks and defenses proposed in the literature, which form the basis of our evaluations and are necessary for understanding our proposed defense mechanism. We only focus on adversarial examples in the domain of image classification, though the same can be crafted for various other computer vision tasks as well.

\subsection{Problem Definition}
\label{sec:problem}
%Before introducing our method, we give a brief overview of the various techniques that have been extensively studied recently for generating adversarial images and defending against them. 
Let $\bm{x}_c \in \mathbb{R}^m$ denote a clean image sample and $\bm{y}_{c}$ its corresponding ground-truth label, where the subscript $\bm{c}$ emphasizes that the image is clean. 
\textit{Untargeted attacks} aim to misclassify a correctly classified example to any incorrect category. 
For these attacks, for a given image classifier $\mathcal{C}: \mathbb{R}^m \rightarrow \{1,2, \cdots, k\}$, an additive perturbation $\bm{\rho} \in  \mathbb{R}^m $ is computed under the constraint that the generated adversarial example $\bm{x}_{adv}= \bm{x}_c + \bm{\rho}$ looks visually similar to the clean image $\bm{x}_c$ i.e., $d(\bm{x}_c, \bm{x}_{adv}) \leq \epsilon$ for some dissimilarity function $d(.,.)$ and the corresponding labels are unequal i.e $\mathcal{C}(\bm{x}_c) \neq \mathcal{C}(\bm{x}_{adv})$. 
\textit{Targeted attacks}  change the correct label to a specified incorrect label, i.e., they seek $\bm{x}_{adv}$ such that $\mathcal{C}(\bm{x}_{adv})= \bm{y}_{tar}$, where $\bm{y}_{tar}$ is a specific class label such that $\bm{y}_{tar} \neq \bm{y}_{c}$. 
An attack is considered successful for an image sample $\bm{x}_c$ if it can find its corresponding adversarial example $\bm{x}_{adv}$ under the given set of constraints. In practice $d(.,.)$ is the $\ell_p$ norm between a clean image and its corresponding adversarial example, where $p \in \{1,\cdots, \infty\}$.

\subsection{Adversarial Attacks}
\label{sec:adversrial}

%Here we give a brief overview of several adversarial attacks in literature which form the basis for our experiments.

\textbf{(a) Fast Gradient Sign Method (FGSM):} This is one of the first attack methods, introduced by Goodfellow~\textit{et al.} \cite{43405}. Given a loss function $L(\bm{x}_c + \bm{\rho},\bm{y}_{c} ; \theta)$, where $\theta$ denotes the network parameters, the goal is to maximize the loss as: 
\begin{equation}
\argmax_{\rho \in \mathcal{R}^m} L(\bm{x}_c + \bm{\rho},\bm{y}_{c} ; \theta).
\end{equation}
% \argmax_{\rho \in \mathcal{\mathbb{R}^m}} f(x)

% $L(\bm{x}_c + \bm{\rho},y_{c} ; \theta)$ \SK{(Please write this as an equation in terms of argmax)}. 
FGSM is a single step attack which aims to find the adversarial perturbations by moving in the opposite direction to the gradient of the loss function w.r.t. the image ($\nabla$):
\begin{equation}
 \bm{x}_{adv}=\bm{x}_c + \epsilon . \text{sign}(\nabla (L(\bm{x}_c,\bm{y}_{c} ; \theta)).
\end{equation}
Here $\epsilon$ is the step size, which essentially restricts the $\ell_{\infty}$ norm of the perturbation. 

\textbf{(b) Iterative Fast Gradient Sign Method (I-FGSM)}
 is an iterative variant of FGSM, introduced by Kurakin \textit{et al.}\cite{kurakin2016}. I-FGSM performs the update as follows:
\begin{equation}
\label{eq-ifgsm}
 \bm{x}_{m+1} = \text{clip}_{\epsilon} (\bm{x}_{m} + \alpha . \text{sign}(\nabla (L(\bm{x}_{m},\bm{y}_{c} ; \theta))),
\end{equation}
where $m\in [0,M]$, $\bm{x}_{0} = \bm{x}_c$ and after $M$ iterations, $\bm{x}_{adv} = \bm{x}_{M}$.

\textbf{(c) Momentum Iterative Fast Gradient Sign Method (MI-FGSM)}, proposed by Dong \textit{et al.} \cite{dong2018boosting}, is similar to I-FGSM with the introduction of an additional momentum term which stabilizes the direction of gradient and helps in escaping local maxima. MI-FGSM is defined as follows:
\begin{equation}
\label{eq:mifgsm}
 g_{m+1} = \mu . g_{m} + \frac{\nabla L(\bm{x}_{m},\bm{y}_{c} ; \theta)}{\parallel \nabla (L(\bm{x}_{m},\bm{y}_{c} ; \theta)) \parallel_1}
\end{equation}
\begin{equation}
 \bm{x}_{m+1} = \text{clip}_{\epsilon} (\bm{x}_{m} + \alpha . \text{sign}(g_{m+1})),
\end{equation}
where $\mu$ is the decay factor, $\bm{x}_{0}=\bm{x}_c$ and  $\bm{x}_{adv} = \bm{x}_{M}$ after $M$ iterations.

\textbf{(d) DeepFool} was proposed by Moosavi-Dezfooli \textit{et al.}\cite{moosavi2016deepfool} and aims to minimize the $\ell_2$ norm between a given image and its adversarial counterpart. The attack assumes that a given image resides in a specific class region surrounded by the decision boundaries of the classifier. The algorithm then iteratively projects the image across the decision boundaries, which is of the form of a polyhydron, until the image crosses the boundary and is misclassified. 
% \MH{This description is not quite clear}

\textbf{(e) Carlini and Wagner (C\&W)}\cite{carlini2017towards} is a strong iterative attack that minimizes an auxiliary variable $\zeta$ as follows:
\begin{equation}
\underset{\zeta}{\text{min}} \parallel \frac{1}{2}(\tanh{(\zeta)}+1) - \bm{x}_c \parallel + c. f(\frac{1}{2}(\tanh{\zeta}+1)),
\end{equation}
where $ \frac{1}{2}(\tanh{(\zeta)}+1) - \bm{x}_c $ is the perturbation $\rho$ and $f(.)$ is defined as 
\begin{equation}
f(\bm{x}_{adv})= \max(Z(\bm{x}_{adv})_{\bm{y}_c} - \max \{Z(\bm{x}_{adv})_n : n \neq \bm{y}_{c} \}, -k).
\end{equation}
Here $Z(\bm{x}_{adv})_n$ are the logit values corresponding to a class $n$ and $k$ is the margin parameter.  The C\&W attack works for various $\ell_p$ norms. % but we in our experiments have attacked the images with an $L_2$ penalty term.

\textbf{(f) DI$^2$FGSM and MDI$^2$FGSM} \cite{xie2018improving}: The aforementioned attacks can be grouped into: single-step and iterative attacks. Iterative attacks have a higher success rate under white-box conditions, but they tend to overfit, and generalize poorly across black-box settings.  
%The aforementioned attacks can be grouped into: single-step and iterative attacks. Compared with their single step attacks counterparts, iterative attacks have a significantly higher success rate in white-box conditions, where the attacker has complete knowledge of the network parameters and its structure. This is because iterative attacks tend to overfit the network parameters and find local maxima which leads to poor generalization across black-box settings. Such attacked images can therefore be easily defended on other architectures.
In contrast, single-step attacks generate perturbed images with fairly improved transferability but a lower success rate in white-box conditions. The recently proposed Diverse-Input-Iterative-FGSM (DI$^2$FGSM) and Momentum-Diverse-Input-Iterative-FGSM (MDI$^2$FGSM) \cite{xie2018improving} methods claim to fill in this gap and improve the transferability of iterative attacks. DI$^2$FGSM performs random image re-sizing and padding as image transformation $\tau(.)$, thus creating an augmented set of images, which are then attacked using I-FGSM as:
\begin{equation}
 \bm{x}_{m+1} = \text{clip}_{\epsilon} (\bm{x}_{m} + \alpha . \text{sign}(\nabla (L(\tau(\bm{x}_{m};p),\bm{y}_{c} ; \theta))).
\end{equation}
Here $p$ is the ratio of transformed images to total number of images in the augmented dataset. MDI$^2$FGSM is a variant, which incorporates the momentum term in DI$^2$FGSM to stabilize the direction of gradients. The overall update for MDI$^2$FGSM is similar to MI-FGSM, with Equation \ref{eq:mifgsm} being replaced by:
\begin{equation}
 g_{m+1} = \mu . g_{m} + \frac{\nabla L(\tau(\bm{x}_{m};p),\bm{y}_{c} ; \theta)}{\parallel \nabla (L(\tau(\bm{x}_{m};p),\bm{y}_{c} ; \theta)) \parallel_1}.
\end{equation}

\begin{SCfigure*}[][h]
  \centering
  \caption{\small{Super-resolution as a Defense Against Adversarial Attacks: The figure illustrates mapping of a sample image from low-resolution to its high-resolution manifold. Adversarial images, which otherwise lie off the manifold of natural images, are mapped in the same domain as the clean natural images, thereby recovering their corresponding true labels. \emph{(Best seen in color and enlarged)} }}
  \includegraphics[width=0.8\textwidth]{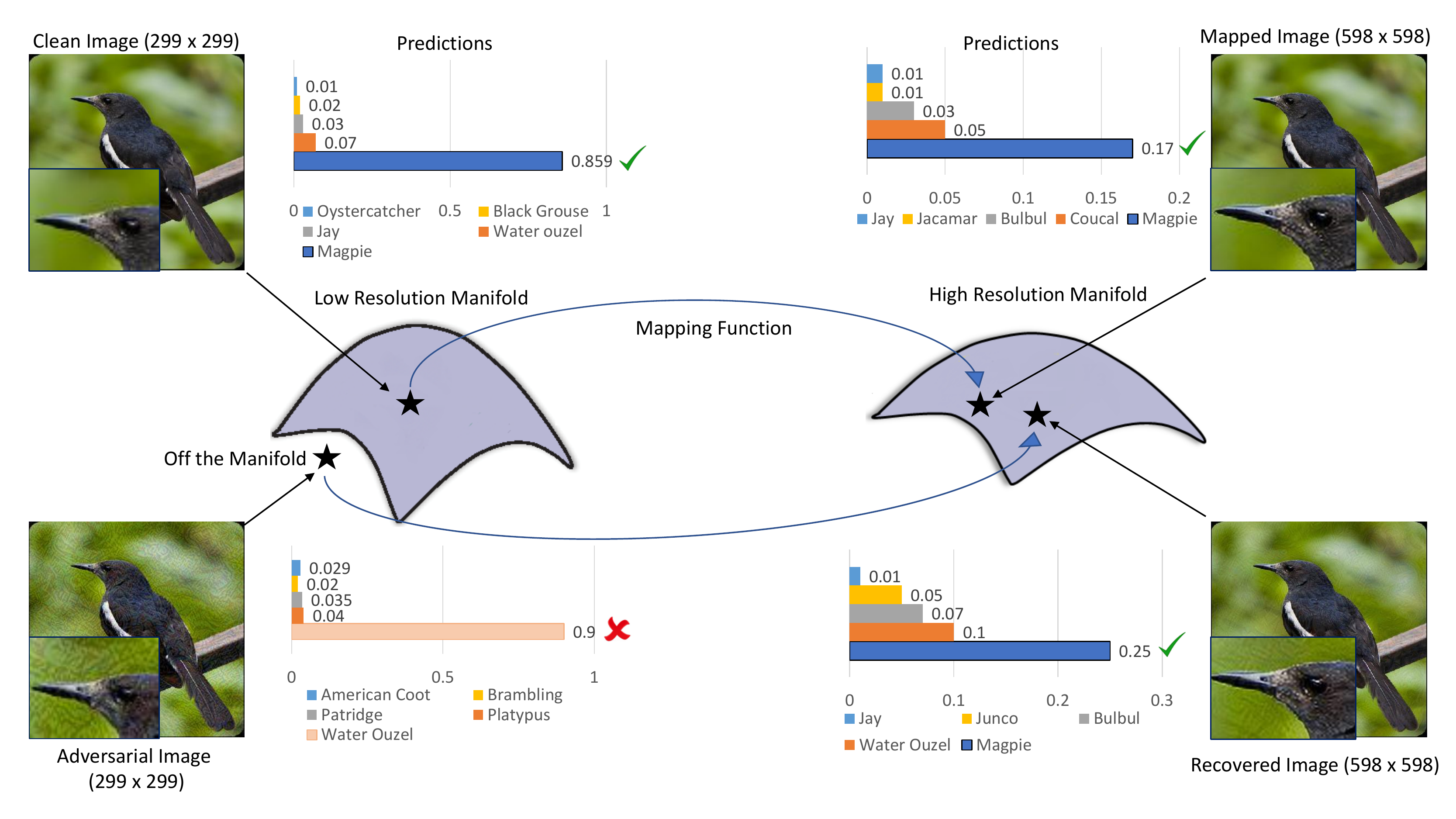}
  \label{fig:mapping}
\end{SCfigure*}

\subsection{Adversarial Defenses}
\label{sec:Adversarial Defenses}
Tremer \textit{et al.}\cite{tramer2017ensemble} proposed ensemble adversarial training, which results in regularizing the network by softening the decision boundaries, thereby encompassing nearby adversarial images. Defensive distillation \cite{papernot2016distillation} improves the model robustness in an essentially similar fashion by retraining a given model using soft labels acquired by a distillation mechanism \cite{hinton2015distilling}. Kurakin \textit{et al. }\cite{kurakin2016adversarial} augmented a training batch of clean images with their corresponding adversarial images to improve robustness. Moosavi-Dezfooli \textit{et al.}\cite{moosavi2017universal}, however, showed that adversarial examples can also be generated for an already adversarially trained model. 

Recently, some defense methods have been proposed in input image transformation domain. Data compression (JPEG image compression) as a defense was studied by \cite{dziugaite2016study,das2017keeping}. JPEG compression deploys a discrete cosine transform to suppress the human-imperceptible, high frequency noise components. Guo \textit{et al.} \cite{guo2017countering}, however, noted that JPEG compression alone is far from being an effective defense. They proposed image transformations using quilting and Total Variance Minimization (TVM). Feature squeezing \cite{xu2017feature} reduces the image resolution either by using bit depth reduction or smoothing filters to limit the adversarial space. A foveation based method was proposed by Luo \textit{et al.}\cite{luo2015foveation}, which shows robustness against weak attacks like L-BFGS\cite{szegedy2013intriguing} and FGSM\cite{43405}. Another closely related work to ours is that of Prakash \textit{et al.}\cite{prakash2018deflecting}, which deflects attention by carefully corrupting less critical image pixels. This introduces new artifacts which reduce the image quality and can result in misclassification. To handle such artifacts, BayesShrink denoising in the wavelet domain is used. It has been shown that denoising in the wavelet domain yields superior performance than other techniques such as bilateral, an-isotropic, TVM and Wiener-Hunt de-convolution \cite{prakash2018deflecting}. Another closely related work is that of Xie \textit{et al.} \cite{xie2017mitigating}, which performs image transformations by randomly re-sizing and padding an image before passing it through a CNN classifier. Xie \textit{et al.} \cite{xie2018feature} showed that adding adversarial patterns to a clean image results in noisy activation maps. A defense mechanism was proposed to perform feature denoising using non-local means, which requires retraining the model end-to-end with adversarial data augmentation. One of the main shortcomings of the aforementioned defense techniques (JPEG compression, PD and foveation based method) is that the transformations degrade the image quality, which results in a loss of significant information from images. %This can be seen in their performance loss on clean image classification.

\section{Proposed Perturbed Image Restoration}
\label{sec:purification}
Existing defense mechanisms against adversarial attacks aim to reduce the effects of added perturbations so as to recover the correct image class. Defenses are being developed along two main directions: \textit{(i)} modifying the image classifier $\mathcal{C}(.)$ to $\mathcal{C}^{'}(.)$ such that it recovers the true label for an adversarial example, i.e. $\mathcal{C}^{'}(\bm{x}_{adv}) = \mathcal{C}(\bm{x}_c) = \bm{y}_{c}$; and \textit{(ii)} transforming the input image such that $\mathcal{C}(\bm{x}_c) = \mathcal{C}(T(\bm{x}_{adv})) = \bm{y}_{c}$, where $T(.)$ is an image transformation function. Ideally, $T(.)$ should be model-agnostic, complex and a non-differentiable function, making it harder for the adversary to circumvent the transformed model by back-propagating the classifier error through it. 

Our proposed approach, detailed below, falls under the second category of defense mechanisms. We propose to use image restoration techniques to purify perturbed images. The proposed approach has two components, which together form a non-differentiable pipeline that is difficult to bypass. As an initial step, we apply wavelet denoising to suppress any noise patterns. The central component of our approach is the super-resolution operation, which enhances the pixel resolution while simultaneously removing adversarial patterns. Our experiments show that image super-resolution alone is sufficient to reinstate classifier beliefs towards correct categories; however, the second step provides added robustness since it is a non-differentiable denoising operation. 

In the following section, we first explain the super-resolution approach (Sec.~\ref{sec:Upsampling}) followed by a description of denoising method (Sec.~\ref{sec:wavelet}). Finally, we summarize the defense scheme in Sec.~\ref{sec:Algorithmic-Description}.

\begin{SCfigure*}[][t]
\includegraphics[clip=true, trim=2cm 1cm 1cm 0.8cm,width = .78\textwidth]{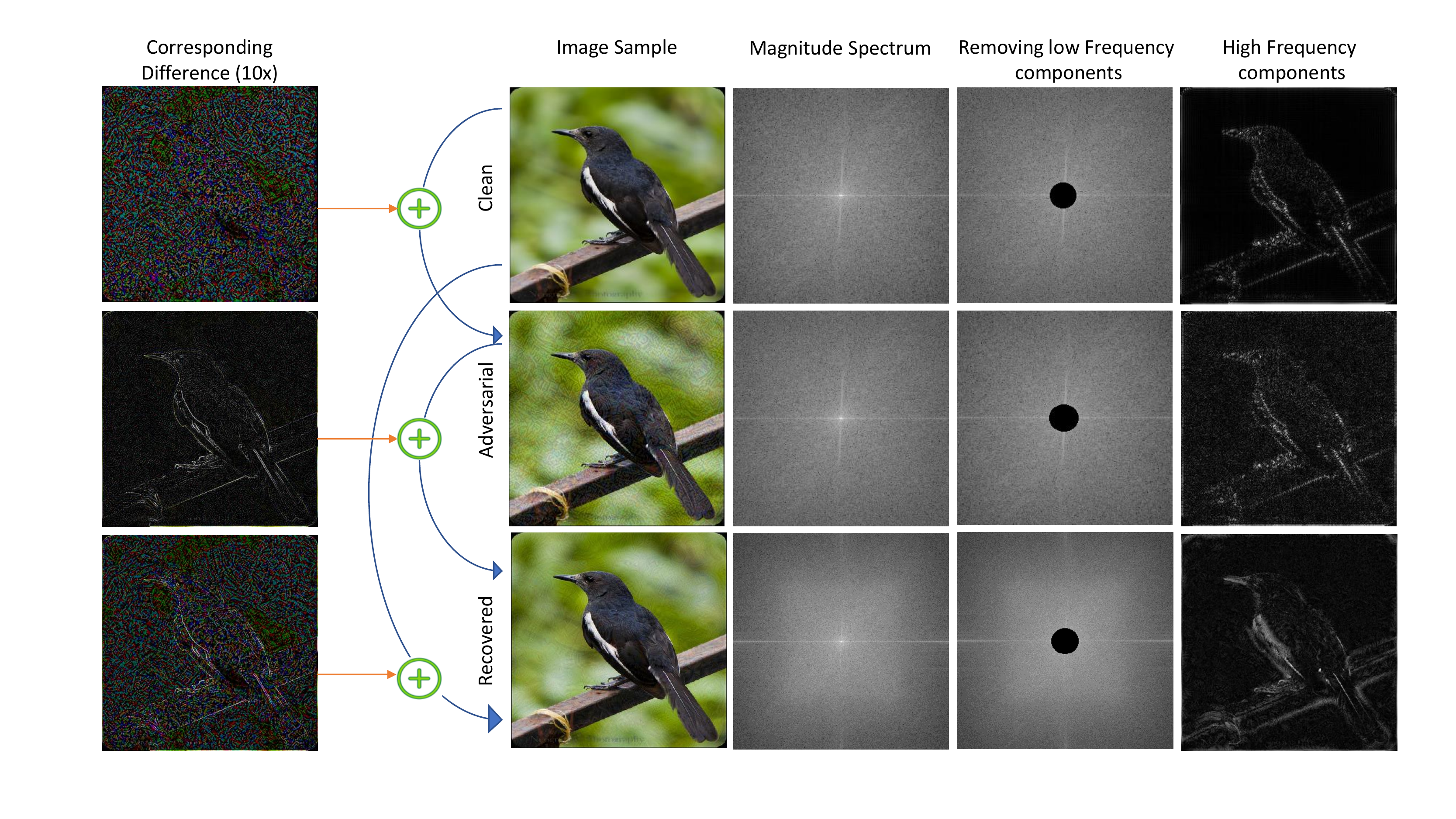}

\caption{\small{ Effect of super-resolution on the frequency distribution of a sample image. A magnitude spectrum for each image is generated using discrete cosine transform (DCT). After removing low-frequency components from the image spectrum (i.e. high pass filtering) inverse DCT is used to visualize the high-frequency components. The IDCT of recovered image shows selective high-frequency components that are added by image super-resolution. The adversarial perturbations were produced using MDI$^2$FGSM with $\epsilon = 16$. }}
% First column is for clean image, second for FGSM ($\epsilon = 10$) attacked image and third for defended image. First row shows a sample image from the dataset. Second row shows the magnitude spectrum after applying discrete fourier transform to the images. Third row shows removing low frequency components which correspond to the centre of magnitude spectrum. The last row shows the output after removing low frequency components and applying inverse fourier transform. \MH{what are the take-away messages from the figure?}}
\label{fig:filtering}
\end{SCfigure*}

\begin{figure}[]
  \centering
  
  \includegraphics[trim={0cm 6.75cm 9.25cm 0cm}, clip, width=0.5\textwidth]{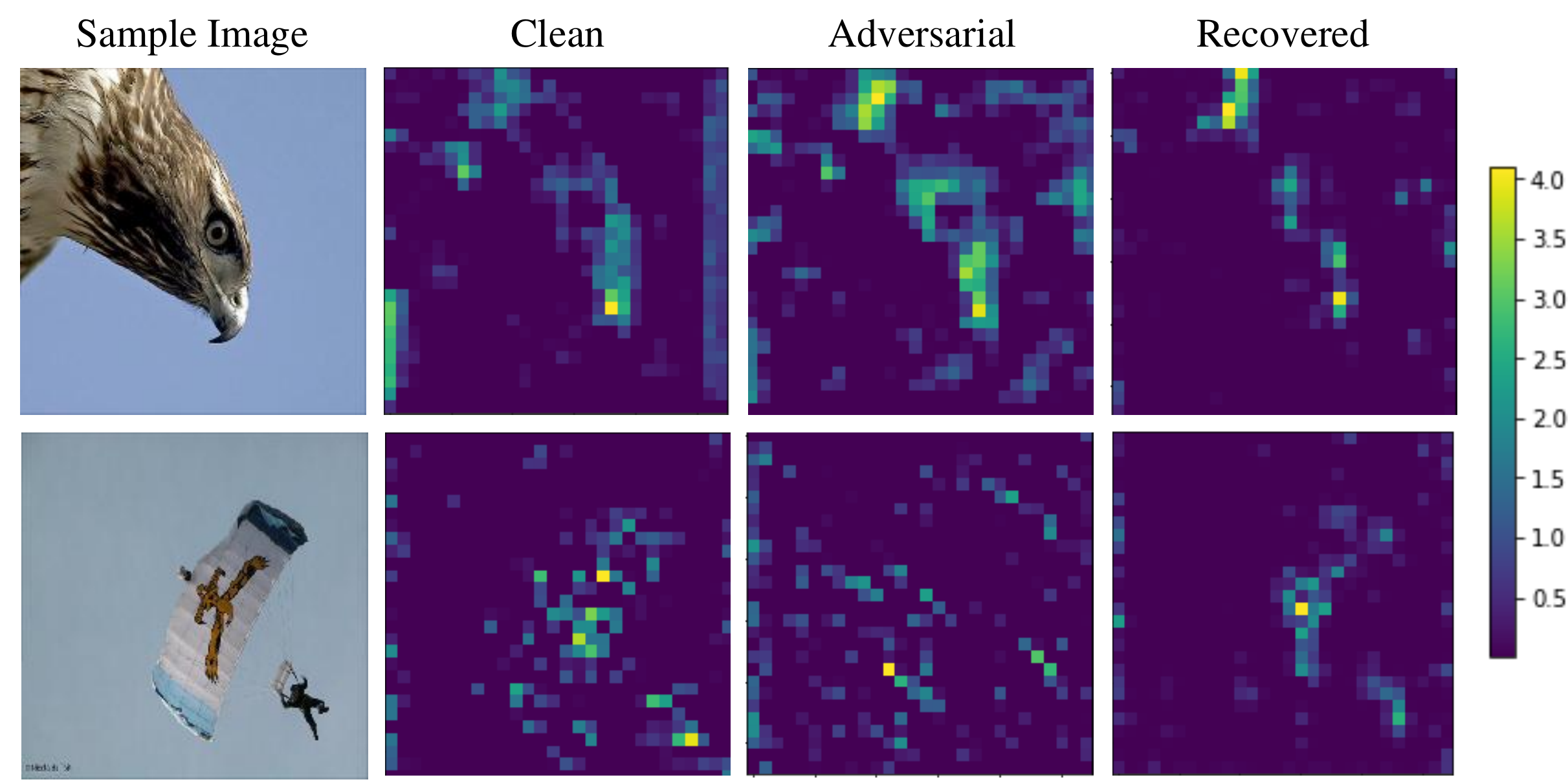}
  \caption{\small{Feature map in the res$_3$ block of an ImageNet-trained ResNet-50 for a clean image, its adversarial counterpart and the recovered image. The adversarial perturbation was produced using FGSM with $\epsilon = 10$. Image super-resolution essentially nullifies the effect of adversarial patterns added by the adversary.}}
  \label{fig:feature_maps}
\end{figure}

\subsection{Super Resolution as a Defense Mechanism}
\label{sec:Upsampling}
% Mapping Function or Purification Network
Our goal is to defend a classification model $\mathcal{C}(\cdot)$ against the perturbed images $\bm{x}_{adv}$ generated by an adversary. Our approach is motivated by the manifold assumption \cite{zhu2009introduction}, which postulates that natural images lie on low-dimensional manifolds. This explains why low-dimensional deep feature representations can accurately capture the structure of real datasets. The perturbed images are known to lie off the low-dimensional manifold of natural images, which is approximated by deep networks \cite{grosse2017statistical}. Gong \textit{et al.} in \cite{gong2017adversarial} showed that a simple binary classifier can successfully separate \textit{off-the-manifold} adversarial images from clean ones and thereby concluded that adversarial and clean data are not twins, despite appearing visually identical. Fig. \ref{fig:mapping} shows a low-dimensional manifold of natural images. Data points from a real-world dataset (say ImageNet) are sampled from a distribution of natural images and can be considered to lie \textit{on-the-manifold}. Such images are referred to as \textit{in-domain} \cite{malinin2018prior}. Corrupting these \textit{in-domain} images by adding adversarial noise takes the images \textit{off-the-manifold}. A model that learns to yield images lying \textit{on-the-manifold} from \textit{off-the-manifold} images can go a long way in detecting and defending against adversarial attacks. We propose to use image super-resolution as a mapping function to remap \textit{off-the-manifold} adversarial samples onto the natural image manifold and validate our proposal through experimentation (see Sec.~\ref{exp-manifold-assumption}). In this manner, robustness  against adversarial perturbations is achieved by enhancing the visual quality of images. This approach provides remarkable benefits over other defense mechanisms that truncate critical information to achieve robustness. 

\textbf{Super-resolution Network:} A required characteristic for defense mechanisms is the ability to suppress fraudulent perturbations added by an adversary. Since these perturbations are generally high-frequency details, we use a super-resolution network that explicitly uses residual learning to focus on such details. These details are added to the low-resolution inputs in each residual block to eventually generate a high-quality, super-resolved image. \A{The network considered in this work is the Enhanced Deep Super-Resolution (EDSR) \cite{lim2017enhanced} network (trained on the DIVerse 2K resolution image (DIV2K) dataset \cite{Agustsson2017NTIRE2C}), which uses a hierarchy of such residual blocks.} While our proposed approach achieves competitive performance with other super-resolution and up-sampling techniques, we demonstrate the added efficacy of using residual learning based EDSR model through extensive experiments (see Sec.~\ref{sec:experiments}). 

\textbf{Effect on Spectral Distribution:}
The underlying assumption of our method is that deep super-resolution networks learn a mapping function that is generic enough to map the perturbed image onto the manifold of its corresponding class images. This mapping function learned with deep CNNs basically models the distribution of real non-perturbed image data. We validate this assumption by analyzing the  frequency-domain spectrum of the clean, adversarial and recovered images in Fig.~\ref{fig:filtering}. It can be observed that adversarial image contains high frequency patterns and the super-resolution operation further injects high frequency patterns to the recovered image. This achieves two major benefits: first, the newly added high-frequency patterns smooth the frequency response of the image (column 5, Fig.~\ref{fig:filtering}) and, second, the super-resolution destroys the adversarial patterns that seek to fool the model.

\A{Fig.~\ref{fig:filtering} also shows that the super-resolved image maintains the high-frequency details close to the original (clean) input image. Still, it is quite different from the clean image, e.g., compare the magnitude spectrum at higher frequencies for the recovered and clean image, which shows a much smoother spread of frequencies in the recovered image. The exact difference between clean and recovered images is shown in the bottom left corner of Fig.~\ref{fig:filtering}, which illustrates the fact that the recovered image is relatively cleaner but has more high-frequency details compared to the original image. Comparing the original noise signal (top left corner in Fig.~\ref{fig:filtering}) and the left-over noise in the recovered image (bottom left corner in Fig.~\ref{fig:filtering}), we can observe that the SR network discards most of the noisy perturbations; however a sparse trace of noise is still present. Also, the SR network reinforces the high-frequency changes along the salient boundaries in the image (notice the response along the bird boundaries). }

\textbf{Effect of Adversarial Perturbations on Feature Maps:}
Adversarial attacks add small perturbations to images, which are often imperceptible to the human eye or generally perceived as small noise in an image in the pixel space. However, this adversarial noise amplifies in the feature maps of a convolutional network, leading to substantial noise \cite{xie2018feature}. Fig.~\ref{fig:feature_maps} shows the feature maps for three clean images, their adversarial counterparts and the defended images chosen from the ResNet-50 res$_3$ block after the activation layer. Each feature map is of $28 \times 28$ dimensions. The features for a clean image sample are activated only at semantically significant regions of the image, whereas those for its adversarial counterpart seem to be focused at semantically irrelevant regions as well. Xie \textit{et al} \cite{xie2018feature} performed feature denoising using non-local means \cite{buades2005non} to improve the robustness of convolutional networks. Their model is trained end-to-end on adversarially perturbed images. Our defense technique recovers the feature maps (Cols 2 and 4, Fig.~\ref{fig:feature_maps}) without requiring any model retraining or adversarial image data augmentation.

\textbf{Advantages of Proposed Method:}
Our proposed method offers a number of advantages. (a) The proposed approach is agnostic to the attack algorithm and the attacked model. (b) Unlike many recently proposed techniques, which degrade critical image information as part of their defense, our proposed method improves image quality while simultaneously providing a strong defense. (c) The proposed method does not require any learning and only uses a fixed set of parameters to purify input images. (d) It does not hamper the classifier's performance on clean images. (e) Due to its modular nature, the proposed approach can be used as a pre-processing step in existing deep networks. Furthermore, our purification approach is equally applicable to other computer vision tasks beyond classification, such as segmentation and object detection. % It will be great to show one small experiment for this.

\subsection{Wavelet Denoising}
\label{sec:wavelet}
Since all adversarial attacks add noise to an image in the form of well-crafted perturbations, an efficient image denoising technique can go a long way in mitigating the effect of these perturbations, if not removing them altogether. Image denoising in the spatial or frequency domain causes a loss of textural details, which is detrimental to our goal of achieving clean image-like performance on denoised images. Denosing in the wavelet domain has gained popularity in recent works. It yields better results than various other techniques including bilateral, anisotropic, Total Variance Minimization (TVM) and Wiener-Hunt de-convolution \cite{prakash2018deflecting}. The main principle behind wavelet shrinkage is that Discrete Wavelet Transform (DWT) of real world signals is sparse in nature. This can be exploited to our advantage since the ImageNet dataset \cite{deng2009imagenet} contains images that capture real-world scenes and objects. Consider an adversarial example $ \bm{x}_{adv}= \bm{x}_{c} + \bm{\rho} $; the wavelet transform of $ \bm{x}_{adv}$ is a linear combination of the wavelet transform of the clean image and noise. Unlike image smoothing, which removes the higher frequency components in an image, DWTs of real world images have large coefficients corresponding to significant image features and noise can be removed by applying a threshold on the smaller coefficients.

\subsubsection{Thresholding}
The thresholding parameter determines how efficiently we shrink the wavelet coefficients and remove adversarial noise from an image. In practice, two types of thresholding methods are used: a) Hard thresholding and b) Soft thresholding. Hard thresholding is basically a non-linear technique, where each coefficient $(\hat{\bm{x}})$ is individually compared to a threshold value $(t)$, as follows:
\[  D(\hat{\bm{x}},t)=
\left\{
	\begin{array}{ll}
		\hat{\bm{x}}  & \mbox{if } |\hat{\bm{x}}| \geq t \\
		0 & \mbox{otherwise}.
	\end{array}
\right.
\]
Reducing the small noisy coefficients to zero and then carrying out an inverse wavelet transform produces an image which retains critical information and suppresses the noise. Unlike hard thresholding where the coefficients larger than $t$ are fully retained, soft thresholding modifies the coefficients as follows:
\[ D(\hat{\bm{x}},t) = \max (0, 1- \frac{t}{|\hat{\bm{x}}|}) \hat{\bm{x}}.
\]
In our method, we use soft-thresholding as it reduces abrupt sharp changes that otherwise occur in hard thresholding. Also, hard- hresholding over-smooths an image, which reduces the classification accuracy on clean non-adversarial images.

Choosing an optimal threshold value $t$ is the underlying challenge in wavelet denoising. A very large threshold value means ignoring larger wavelets, which results in an over-smoothed image. In contrast, a small threshold allows even the noisy wavelets to pass, thus failing to produce a denoised image after reconstruction. Universal thresholding is employed in \textit{VisuShrink} \cite{donoho1994ideal} to determine the threshold parameter $t_{vs}$ for an image $X$ with $n$ pixels as $t_{vs}= \sigma_{\rho} \sqrt{2\ln{(n)}}$, where $\sigma_{\rho}$ is an estimate of the noise level. \textit{BayesShrink} \cite{chang2000adaptive} is an efficient method for wavelet shrinkage which employs different thresholds for each wavelet sub-band by considering Gaussian noise. Suppose $\hat{\bm{x}}_{adv}= \hat{\bm{x}}_c + \hat{\bm{\rho}}$ is the wavelet transform of an adversarial image, since $\hat{\bm{x}}_c$ and $\hat{\bm{\rho}}$ are mutually independent, the variances $\sigma^2_{\bm{x}_{adv}}$, $\sigma^2_{\bm{x}_{c}}$ and $\sigma^2_{\rho}$ of $\hat{\bm{x}}_{adv}$, $\hat{\bm{x}}_c$, $\hat{\bm{\rho}}$, respectively, follow: $ \sigma^2_{\bm{x}_{adv}}= \sigma^2_{\bm{x}_{c}} + \sigma^2_{\rho}$. A wavelet sub-band variance for an adversarial image is estimated as:
\[\sigma^2_{\bm{x}_{adv}}= \frac{1}{M} \sum_{m=1}^{M} W^2_m ,\] where $W^2_m$ are the sub-band wavelets and $M$ is the total number of wavelet coefficients in a sub-band. The threshold value for \textit{BayesShrink} soft-thresholding is given as:
\[  t_{bs}=
\left\{
	\begin{array}{ll}
		{\sigma^2_{\rho}}/{\sigma_{\bm{x}_{c}}} & \mbox{if } \sigma^2_{\rho} < \sigma^2_{\bm{x}_{adv}} \\
		\max(|W_m|) & \mbox{otherwise}.
	\end{array}
\right.
\]
In our experiments, we explore both \textit{VisuShrink} and \textit{BayesShrink} soft-thresholding and find the latter to perform better and provide visually superior denoising.

\subsection{Algorithmic Description}
\label{sec:Algorithmic-Description}

An algorithmic description of our end-to-end defense scheme is provided in Algorithm \ref{algo:b}. We first smooth the effect of adversarial noise using soft wavelet denoising. This is followed by employing super resolution as a mapping function to enhance the visual quality of images. Super resolving an image maps the adversarial examples to the natural image manifold in high-resolution space, which otherwise lie \textit{off-the-manifold} in low-resolution space. The recovered image is then passed through the same pre-trained models on which the adversarial examples were generated. As can be seen, our model-agnostic image transformation technique is aimed at minimizing the effect of adversarial perturbations in the image domain, with little performance loss on clean images. Our technique causes minimal depreciation in the classification accuracy of non-adversarial images.% as seen in Table \ref{table:clean_imagenet}.

 % \begin{table}[h]
% % \caption{Choosing Hyper-parameter $S$. Top-1 classification accuracy of Inception v-3 model.}
% % \label{table:algorithm1}
% \centering
% \begin{center}
% \begin{tabular}{l}
% \hline 
% \multicolumn{1}{c}{\textbf{Algorithm 1} Wavelet Denoising }\\
% % \textbf{Algorithm 1} Wavelet Denoising \\
% \hline
% \textbf{input:} Corrupted image $\bm{x}_{adv} = \bm{x}_c + \bm{\rho}$\\
% \textbf{output:} Denoised image $\bm{x}^{'} = D(\bm{x}_{adv}) $ \\
% 1. Convert the RGB image to $YC_{b}C_{r}$ color space, where $Y$ \\and $C_b,C_r$ 
% represent luminance and chrominance respectively.\\
% 2. Convert the image to wavelet domain $\hat{X}_{adv}= \hat{X_c}+ \hat{\rho}$ \\using discrete wavelet transform.\\
% 3. Remove noisy wavelet coefficients using BayesShrink\\ soft-thresholding.\\
% 4. Invert the shrunken wavelet coefficients using Inverse\\ Wavelet Transform (IWT).\\
% 5. Revert the image back to RGB.\\
% \hline
% \end{tabular}
% \end{center}
% \end{table}

\begin{algorithm}[h]

% \centering
%\DontPrintSemicolon % Some LaTeX compilers require you to use \dontprintsemicolon    instead
%  Step 1: Wavelet Denoising \\
\footnotesize{\tcc{Image Denoising}}
\footnotesize{\KwIn{Corrupted image $\bm{x}_{adv} = \bm{x}_c + \bm{\rho}$}}
\footnotesize{\KwOut{Denoised image $\bm{x}^{'} = D(\bm{x}_{adv}) $}}
\vspace{3mm}
\footnotesize{ Convert the RGB image to $YC_{b}C_{r}$ color space, where $Y$ and $C_b,C_r$ 
represent luminance and chrominance respectively.\\

Convert the image to wavelet domain $\hat{X}_{adv}= \hat{X_c}+ \hat{\rho}$ using discrete wavelet transform.\\

Remove noisy wavelet coefficients using BayesShrink soft-thresholding.\\ 
Invert the shrunken wavelet coefficients using Inverse Wavelet Transform (IWT).\\
Revert the image back to RGB.}\\ 

\vspace{3mm}
\tcc{Image Super-Resolution}
\footnotesize{\KwIn{Denoised image $\bm{x}^{'} = D(\bm{x}_{adv}) $}}
\footnotesize{\KwOut{Super Resolved Image $\bm{x}_{t} = M(\bm{x}^{'}) $ }}
\vspace{3mm}

Map adversarial samples back to natural image manifold using deep super resolution network: $M(\cdot)$.\\
Forward the recovered images to the attacked model for correct prediction. \\
%Testing top-1 model accuracy on recovered images.\\

% Step 2: Image Super Resolution \\
\caption{\footnotesize{ Defending Against Adversarial Attacks with Image Restoration (Wavelet Denoising + Super Resolution)}}
\label{algo:b}
\end{algorithm}

% \begin{enumerate}
% \item Perform multi-scale decomposition to convert the adversarial image to $ YC_{b}C_{r}$ space and then in wavelet domain using discrete wavelet transform.
% \item Remove noisy wavelet coefficients using BayesShrink soft-thresholding. 
% \item Invert the shrunken wavelet coefficients using Inverse Wavelet Transform (IWT).
% \item Revert the image back to RGB.
% \end{enumerate}

% Before passing through a convolutional neural network the resulting denoised image is super resolved to a higher resolution as:

% \begin{enumerate}
% \item
% \end{enumerate}

\begin{table*}[]
\caption{Performance comparison with state-of-the art defense mechanisms on 5000 images from ILSVRC validation set. The images are selected such that the respective classifier achieves $100\%$ accuracy. Our proposed defense consistently achieves superior performance across three different models and various adversarial attacks.}
\label{table-comparison}

\centering
\begin{center}
\resizebox{.99\textwidth}{!}{
\begin{tabular}{ccccccccccc}
\hline 
%&&&\\
%\multicolumn{5}{c}{\textbf{Inception v-3 model}}\\
%\hline \hline
%&&&\\
\rowcolor{black!10} \textbf{Model} &\textbf{Clean Images} & \textbf{FGSM-2} &\textbf{FGSM-5}& \textbf{FGSM-10}  & \textbf{I-FGSM}  &\textbf{DeepFool} &  \textbf{C\&W}  & \textbf{MI-FGSM} 
 &  \textbf{DI$^2$FGSM}  & \textbf{MDI$^2$FGSM} \\
\hline \hline
\multicolumn{11}{c}{No Defense}\\
\hline
Inception v-3 &100&31.7&28.7&30.5&11.4 & 0.4&0.8& 1.7& 1.4&0.6\\
ResNet-50 &100&12.2&7.0&6.1&3.4 & 1.0& 0.1& 0.4& 0.3 & 0.2\\
Inception ResNet v-2 &100&59.4&55.0&53.6&21.6 & 0.1&0.3& 0.5& 1.5& 0.6\\
\hline
\multicolumn{11}{c}{JPEG Compression (Das \textit{et al.} \cite{das2018shield})}\\
\hline
Inception v-3& 96.0&62.3& 54.7 & 48.8& 77.5& 81.2& 80.5& 69.4 & 2.1& 1.3\\
ResNet-50 & 92.8 &57.6 & 49.0& 42.9& 74.8& 77.3& 81.3& 70.8& 0.7&0.4\\
Inception ResNet v-2 & 95.5 &67.0& 55.3& 53.7& 81.3& 83.9& 83.1& 72.8& 1.6& 1.1\\
\hline
\multicolumn{11}{c}{Random resizing + zero padding (Xie \textit{et al.} \cite{xie2017mitigating})}\\
\hline
Inception v-3 &\textbf{97.3} &69.2&57.3&53.2& 90.6 &88.9&89.5&89.5&7.0&5.8\\
ResNet-50 &92.5 &66.8& 55.7 & 48.8 & 88.2 & 90.9& 87.5 & 88.0 & 6.6 & 4.2\\
Inception ResNet v-2 &\textbf{98.7} &70.7&59.1&55.8& 87.5&89.7&88.0&88.3&7.5&5.3\\
\hline
\multicolumn{11}{c}{Quilting + Total Variance Minimization (Guo \textit{et al.} \cite{guo2017countering})}\\
\hline
Inception v-3 & 96.2& 70.2& 62.0& 54.6 &85.7 &85.9&85.3&84.5&4.1&1.7\\
ResNet-50 &93.1& 69.7&61.0& 53.3 & 85.4& 85.0 & 84.6& 83.8& 3.6& 1.1\\
Inception ResNet v-2 & 95.6&74.6& 67.3&59.0 & 86.5& 86.2& 85.3&84.8 & 4.5 & 1.2\\
\hline
\multicolumn{11}{c}{Pixel Deflection (Prakash \textit{et al.} \cite{prakash2018deflecting})}\\
\hline
Inception v-3&91.9 &71.1&66.7&58.9& 90.9 & 88.1&90.4&90.1&57.6&21.9\\
ResNet-50 & 92.7&84.6 & 77.0 &\textbf{66.8} &91.2 &90.3& 91.7 & 89.6& 57.0 & 29.5\\
Inception ResNet v-2& 92.1 &78.2&75.7&71.6& 91.3 &88.9&89.7&89.8&57.9&24.6\\
\hline
\multicolumn{11}{c}{Our work: Wavelet Denoising + Image Super Resolution}\\
\hline
Inception v-3 &97.0&\textbf{94.2}&\textbf{87.9}&\textbf{79.7}&\textbf{96.2} &\textbf{96.1}&\textbf{96.0}&\textbf{95.9}&\textbf{67.9}&\textbf{31.7}\\
ResNet-50 & \textbf{93.9}&\textbf{86.1} &\textbf{77.2} & 64.9 &\textbf{92.3} & \textbf{91.5 }&\textbf{93.1} &\textbf{92.0} &\textbf{60.7} & \textbf{31.9}\\
Inception ResNet v-2& 98.2&\textbf{95.3}&\textbf{87.4}&\textbf{82.3}&\textbf{95.8} & \textbf{96.0}&\textbf{95.6}&\textbf{95.0}&\textbf{69.8}&\textbf{35.6}\\
\hline
\end{tabular}}
\end{center}
\end{table*}

\section{Experiments}
\label{sec:experiments}

%\SK{Can we also include some experiments with generic up-sampling or down+upsampling and any comparisons of that with our final SR based approach? }
%\A{done. Table VII showing performance of Nearest neighbor, Bilinear and Bicubic}

\noindent \textbf{Models and Datasets: }We evaluate our proposed defense and compare it with existing methods for three different classifiers: Inception-v3, ResNet-50 and InceptionResNet v-2. For these models, we obtain ImageNet pre-trained weights from TensorFlow's GitHub repository \footnote{https://github.com/tensorflow/models/tree/master/research/slim}, and do not perform any re-training or fine-tuning. The evaluations are done on a subset of 5000 images from the ILSVRC \cite{deng2009imagenet} validation set. The images are selected such that the respective model achieves a top-1 accuracy of $100\%$ on the clean non-attacked images. Evaluating defense mechanisms on already misclassified images is not meaningful, since an attack on a misclassified image is considered successful as per the definition. We also perform experiments on the NIPS 2017 Competition on Adversarial Attacks and Defenses DEV dataset \cite{kurakin2018adversarial}. The dataset is collected by Google Brain organizers, and consists of $1000$ images of size $299 \times 299$. An ImageNet pre-trained Inception v-3 model achieves $95.9\%$ top-1 accuracy on NIPS 2017 DEV images. \\

\noindent \textbf{Attacks: } We generate attacked images using different techniques, including Fast Gradient Sign Method (FGSM) \cite{43405}, iterative FGSM (I-FGSM) \cite{kurakin2016}, Momentum Iterative FGSM (MI-FGSM) \cite{dong2018boosting}, DeepFool \cite{moosavi2016deepfool}, Carlini and Wagner \cite{carlini2017towards}, Diverse Input Iterative FGSM (DI$^2$FGSM) and Momentum Diverse Input Iterative FGSM (MDI$^2$FGSM) \cite{xie2018improving}. We use publicly available implementations of these methods: Cleverhans\cite{papernot2018cleverhans}, Foolbox\cite{rauber2017foolbox} and codes\footnote{https://github.com/dongyp13/Non-Targeted-Adversarial-Attacks} \footnote{https://github.com/cihangxie/DI-2-FGSM} provided by \cite{dong2018boosting,xie2018improving}. For FGSM, we generate attacked images with $\epsilon \in \{2,5,10\}$ and for iterative attacks, the maximum perturbation size is restricted to $16$. \A{All the adversarial images are generated for the undefended models, after which various defense schemes are implemented in \textit{gray-box} settings.}\\

\noindent \textbf{Defenses: } \A{We compare our proposed defense with a number of recently introduced state-of-the-art image transformation based defense schemes in the literature.} These include JPEG Compression \cite{das2018shield}, Random Resizing and Padding \cite{xie2017mitigating}, Image quilting + total variance minimization \cite{guo2017countering} and Pixel Deflection (PD)\cite{prakash2018deflecting}. We use publicly available implementations \footnote{https://github.com/poloclub/jpeg-defense}  \footnote{https://github.com/cihangxie/NIPS2017\_adv\_challenge\_defense}
\footnote{https://github.com/facebookresearch/adversarial\_image\_defenses}
\footnote{https://github.com/iamaaditya/pixel-deflection} of these methods. All experiments are run on the same set of images and against the same attacks for a fair comparison. 

For our experiments, we explore two broad categories of Single Image Super Resolution (SISR) techniques: \textit{i)} Interpolation based methods and \textit{ii)} Deep Learning (DL) based methods. Interpolation based methods like Nearest Neighbor (NN), Bi-Linear and Bi-cubic upsampling are computationally efficient, but not quite robust against stronger attacks (DI$^2$FGSM and MDI$^2$FGSM). Recently proposed DL based methods have shown superior performance in terms of Peak Signal to Noise Ratio (PSNR) and Structural Similarity Index (SSIM), and the mean squared error (MSE). Here, we consider three DL based SISR techniques, \textit{i)} Super Resolution using ResNet model (SR-ResNet) \cite{ledig2017photo}, \textit{ii)} Enhanced Deep Residual Network  for SISR (EDSR) \cite{lim2017enhanced} and \textit{iii)} Super Resolution using Generative Adversarial Networks (SR-GAN) \cite{ledig2017photo}. Our experiments show that EDSR consistently performs better. EDSR builds on a residual learning \cite{he2016deep} scheme that specifically focuses on high-frequency patterns in the images. Compared to the original ResNet, EDSR demonstrates substantial improvements by removing Batch Normalization layers (from each residual block) and ReLU activation (outside residual blocks). 

\subsection{Manifold Assumption Validation}
\label{exp-manifold-assumption}
In this paper we propose that clean and adversarial examples lie on different manifolds and super-resolving an image to a higher dimensional space remaps the adversarial sample back to the natural image manifold.

To validate this assumption, we fine-tune a pre-trained Inception v-3 model on the ImageNet dataset as a binary classifier using 10,000 pairs of clean and adversarial examples (generated from all the aforementioned attack techniques). We re-train the top-2 blocks while freezing the rest with a learning rate reduced by a factor of 10. The global average pooling layer of the model is followed by a batch normalization layer, drop-out layer and two dense layers (1024 and 1 nodes, respectively). Our model efficiently leverages the subtle difference between clean images and their adversarial counterparts and separates the two with a very high accuracy (99.6\%). To further validate our assumption on super-resolution, we test our defended images using this binary classifier. The classifier labels around 91\% of the super-resolved images as clean, confirming that the vast majority of restored samples lie on the natural image manifold.

In Figure.~\ref{fig:manifold-assumption}, we plot the features extracted from the last layer of the binary classifier to visualize our manifold assumption validation. We reduce the dimensionality of features to 3 for visualization (containing 90\% of variance) using Principle Component Analysis.

\begin{table*}[]
\caption{Top-1 accuracy comparison for different defense mechanisms on NIPS-DEV dataset on Inception v-3 model.}
\label{table-comparison_nipsdata}
\centering
%\begin{center}
\begin{adjustbox}{max width=0.99\textwidth}
\begin{tabular}{lccccc|cc}
\hline 
\rowcolor{black!10} \textbf{Attack} & \textbf{No Defense} & \textbf{Das \textit{et al.} \cite{das2018shield}} & \textbf{Xie \textit{et al.} \cite{xie2017mitigating}} & \textbf{Guo \textit{et al.} \cite{guo2017countering} }& \textbf{Prakash \textit{et al.} \cite{prakash2018deflecting}}   & \textbf{Ours (SR)}  & \textbf{Ours (WD + SR)}\\

\hline \hline
Clean  & 95.9 & 89.7& \textbf{92.0} & 88.8 &  86.5   &90.4   & \textbf{90.9}\\

FGSM-2 & 22.1 &58.3&  65.2 & 68.3&   70.7   & 87.1  & \textbf{87.5}\\ 
%\hline 
FGSM-5 & 20.0 & 50.2& 52.7 &58.0 &  62.9 & 79.6   & \textbf{79.9}\\ 
%\hline
FGSM-10 & 23.1 &43.5& 47.5& 50.5 &   54.2  & 69.8 &  \textbf{70.1}\\
%\hline
I-FGSM & 10.1& 75.8  & 85.3 & 80.9& 86.2   & 89.7 &   \textbf{90.1}\\
%\hline
DeepFool & 1.0 & 77.0 & 84.7 & 80.1& 84.2    & 90.2  & \textbf{90.4}\\
%\hline
C\&W & 0.3 & 76.3  & 84.8  & 80.3& 84.9   & 90.5  & \textbf{90.7}\\
%\hline
MI-FGSM & 1.4 & 72.4 & 83.6 & 78.2& 84.0     & 89.4   &\textbf{89.8}\\
%\hline
DI$^2$FGSM & 1.7  & 2.0 & 5.1& 3.1 & 54.6    & 48.9  &\textbf{63.8}\\
%\hline
MDI$^2$FGSM & 0.6& 1.3 & 4.0& 1.8& 20.4    & 26.1   & \textbf{28.7}\\
\hline
\end{tabular}
%\end{center}  \vspace{-3mm}
\end{adjustbox}
\end{table*}

\begin{table}[htp]
\caption{Performance comparison of various super-resolution techniques in the literature. The up-scaling factor $S=2$. Top-1 accuracies are reported. }
\label{table:sr-comparison}
\centering
\begin{center}
\resizebox{1\columnwidth}{!}{
\begin{tabular}{l*{4}{c}r}
%\hline \hline &&&\\
%\multicolumn{5}{c}{\textbf{Comparison of various super-resolution techniques.}}\\
\hline 
\rowcolor{black!10} \textbf{Attack}       & \textbf{No Defense}  &\textbf{SR-ResNet \cite{ledig2017photo}} &  \textbf{SR-GAN \cite{ledig2017photo}} & \textbf{EDSR \cite{lim2017enhanced}}    \\
\hline\hline
Clean            & 100.0 & 94.0 & 92.3  & \textbf{96.2} \\
FGSM-2           & 31.7   &89.5 &  85.7  &   \textbf{92.6}\\
FGSM-5           & 28.7   & 83.7&  80.1  &  \textbf{85.7} \\
FGSM-10          & 30.5  & 69.9 & 69.0  &  \textbf{73.3} \\
I-FGSM           &  11.4 &  93.4    &    91.0   & \textbf{95.9} \\
DeepFool         & 0.4  &  93.2& 93.0    &  \textbf{95.5} \\
C\&W             & 0.8   & 93.3&  91.3 &  \textbf{95.6} \\
MI-FGSM          & 1.7  & 92.6  &   87.6 &  \textbf{95.2} \\
DI$^2$FGSM       & 1.4  & 54.3 &48.9    &  \textbf{57.2} \\
MDI$^2$FGSM      & 0.6  &  24.9 &  23.0 &  \textbf{27.1} \\
\hline 
% &&&\\
\end{tabular}}
\vspace{-1em}
\end{center}
\end{table}

\subsection{Results and Analysis}
\label{subsec:results}

Table \ref{table-comparison} shows the destruction rates of various defense mechanisms on 5000 ILSVRC validation set images. Destruction rate is defined as the ratio of successfully defended images \cite{kurakin2016}. A destruction rate of $100\%$  implies that all images are correctly classified after applying the defense mechanism. It should be noted that we define destruction rates in terms of top-1 classification accuracy, which makes defending against attacks more challenging since we have to recover the exact class label. 
%For a fair comparison with existing methods including JPEG Compression \cite{das2018shield},  Random Resizing + Padding \cite{xie2017mitigating}, Image quilting + total variance minimization \cite{guo2017countering} and Pixel Deflection (PD) \cite{prakash2018deflecting}, we applied their respective image transformations on the same set of images for the three evaluated CNN models. 
`No Defense' in Table~\ref{table-comparison} shows the model performance on generated adversarial images. A lower value under `No Defense' is an indication of a strong attack. The results show that iterative attacks are better at fooling the model compared with the single-step attacks. The iterative attacks, however, are not transferable and are easier to defend. Similarly, targeted attacks are easier to defend compared with their non-targeted counterparts, as they tend to over-fit the attacked model \cite{carlini2017towards}. Considering them as weak attacks, we therefore only report the performance of our defense scheme against more generic non-targeted attacks. 

For the iterative attacks (C\&W and DeepFool), both Random Resizing + Padding and PD achieve similar performance, successfully recovering about 90\% of the images. In comparison, our proposed super-resolution based defense recovers about 96\% of the images. For the single-step attack categories, Random Resizing + Padding fails to defend. This is also noted in \cite{xie2017mitigating}. To overcome this limitation, an ensemble model with adversarial augmentation is used for defense. Compared with the JPEG compression based defense \cite{das2017keeping}, our proposed method achieves a substantial performance gain of 31.1\% for FGSM ($\epsilon = 10$). In the single-step attacks category (e.g., FGSM-10), our defense model outperforms Random Resizing + Padding and PD by a large margin of 26.7\% and 21.0\%, respectively. For the recently proposed strong attack (MDI$^2$FGSM), all defense techniques (JPEG compression, Random Resizing + Padding, Quilting + TVM and PD) largely fail, recovering only 1.3\%, 5.8\%, 1.7\% and 21.9\% of the images, respectively. In comparison, the proposed image super-resolution based defense can successfully recover $31.3\%$ of the images.

\begin{table}[htp]
\caption{Performance of Nearest Neighbor, Bi-linear and Bi-cubic image resizing techniques as a defense. Evaluation is done on NIPS-DEV dataset using a pretrained Inception v-3 model. US: Upsample; DS: Downsample. The US and DS factor is 2.}
\label{table-nn-comparison}
\begin{center}
\resizebox{.98\columnwidth}{!}{
\begin{tabular}{l|ccc|ccc|ccc}
\hline
\rowcolor{black!10}  \multicolumn{1}{l}{\textbf{Transform}}& \multicolumn{3}{c}{\textbf{Nearest Neighbor}} &\multicolumn{3}{c}{\textbf{Bi-linear}} &\multicolumn{3}{c}{\textbf{Bi-cubic}}\\
\hline \hline
% & \multicolumn{3}{c}{No Defense} &\multicolumn{3}{c}{No Defense} &\multicolumn{3}{c}{No Defense} \\
US & \cmark & - & - & \cmark & - & - & \cmark  & - & - \\
US $\rightarrow$ DS & - & \cmark & - & - & \cmark & - & - & \cmark & - \\
DS $\rightarrow$ US & - & - & \cmark & - & - & \cmark & - & - & \cmark  \\
%\textbf{}   & US& US   & DS  & US& US   & DS& US& US   & DS\\
%\textbf{}   &($\times$ 2)& + & +&($\times$ 2)& + & +&($\times$ 2)& + & +\\
%\textbf{}   && DS& US&& DS& US&& DS& US\\
\hline
\textbf{Attack} & \\
\hline
{Clean}   & 94.9 & 93.5&84.1 & 94.3& 91.1& 86.2& 93.9& 89.1& 86.0\\
{FGSM-2}   &74.2& 25.9& 23.8& 73.5&24.0& 21.4& 71.2&20.3& 19.5\\
{FGSM-5}  &61.0&18.6& 18.1& 60.5& 18.1& 17.0& 54.8& 18.7& 18.0\\
{FGSM-10} &52.9& 16.8& 16.0& 50.1& 15.7& 15.4 &49.2& 16.2& 15.8\\
{I-FGSM} &86.4& 45.9& 43.0& 83.4& 41.9& 40.6& 82.1& 37.5& 35.6\\
{DeepFool} &87.3& 43.0& 41.2& 80.6& 40.7& 39.7& 80.1& 34.5& 30.1\\
{C\&W} &82.5& 44.4& 41.2& 80.1& 41.9& 37.6& 79.3&39.6& 36.0\\
{MI-FGSM}  &81.4& 41.0& 38.0& 80.0 & 42.9& 40.3& 80.7& 41.2& 39.8\\
{DI$^2$FGSM}  &36.0& 5.8& 3.9& 34.8& 6.1& 4.9& 31.2& 7.8& 5.6\\
{MDI$^2$FGSM}  &16.1& 3.8& 2.0& 10.2& 3.9& 3.5& 9.6& 2.0& 1.7\\
\hline 
\end{tabular}}
\end{center}
\end{table}

We show a further performance comparison of our proposed defense with other methods on the NIPS-DEV dataset in Table~\ref{table-comparison_nipsdata}. Here, we only report results on Inception v-3, following the standard evaluation protocols as per the competition's guidelines \cite{kurakin2018adversarial}. Inception v-3 is a stronger classifier, and we expect the results to generalize across other classifiers. Our experimental results in Table~\ref{table-comparison_nipsdata} show the superior performance of the proposed method. 

\subsection{Adversarial Training}

\A{\emph{Adversarial Training} has been shown to enhance many recently proposed defense methods \cite{kurakin2018adversarial} under \textit{white-box} attack settings. Taking insights from the effect of super-resolution under \textit{gray-box} settings, we introduce a robust adversarial training paradigm that enhances the performance of traditional adversarial training. For this, we jointly train our model on an augmented dataset comprising of clean, attacked and super-resolved images (for CIFAR-10 dataset) to improve the generalization of adversarial training. Our results in Table~\ref{table-robust-adv-train} indicate that our adversarially trained model provides enhanced robustness against \textit{white-box} attacks. Below we describe our experimental settings used for training and evaluation.}

\A{\textit{Experimental Settings:} The adversarial samples used for the training process are generated using Fast Gradient Sign Method (FGSM) \cite{43405} by uniformly sampling $\epsilon$ from an interval of [0.01, 0.05] for CIFAR-10 dataset. These attacked images are then super-resolved to form the augmented dataset for training. We evaluate the model's robustness against single-step as well as strong iterative attacks in \textit{white-box} conditions. The number of iterations for Iterative Fast Gradient Sign Method (I-FGSM) \cite{kurakin2016adversarial}, Momentum Iterative FGSM (MI-FGSM)  \cite{dong2018boosting} and Projected Gradient Descent (PGD)  \cite{madry2017towards} are set to 10 with a step size of $\epsilon/10$ for I-FGSM and MI-FGSM and $\epsilon/4$ for the PGD attack. The iteration steps for the Carlini \& Wagner (C\&W) attack \cite{carlini2017towards} are 1,000 with a learning rate of 0.01. We used a ResNet-110 model for training the CIFAR-10 dataset. Table~\ref{table:architecture} gives the detailed architecture of the model used. }

\begin{table}[h]
\caption{\small{\A{Comparison of our robust adversarial training method on CIFAR-10 dataset against various \textit{white-box} attacks (numbers show robustness, higher is better). We report results without adversarial training (baseline), with adversarial training (AdvTrain) and adversarial training with training dataset augmented with super-resolved images (Robust AdvTrain). Here $\epsilon$ is the perturbation size and $c$ is the initial constant for C\&W attack. }}}

\label{table-robust-adv-train}
\begin{center}
\resizebox{.5\textwidth}{!}{
\begin{tabular}{c||c|c|c|c}
\hline \hline

\cellcolor{black!10} Attacks & \cellcolor{black!10} Params. &\cellcolor{black!10} Baseline &\cellcolor{black!10} AdvTrain &\cellcolor{black!10} Robust AdvTrain \\

\hline \hline
\multirow{2}{*}{No Attack} & \multirow{2}{*}{-}   & \multirow{2}{*}{\textbf{90.8}} & \multirow{2}{*}{84.5} & \multirow{2}{*}{87.4}\\
 & & & &\\
\hline
\multirow{2}{*}{FGSM}   & $\epsilon = 0.02 $ & 36.5 & 44.3  & \textbf{48.5}\\
 & $\epsilon = 0.04 $ & 19.4 & 31.0 & \textbf{37.1}\\
\hline
\multirow{2}{*}{I-FGSM} &   $\epsilon = 0.01 $ & 26.0 & 32.6 & \textbf{35.3} \\
 & $\epsilon = 0.02 $ & 6.1 & 7.8 & \textbf{10.3}\\
\hline
\multirow{2}{*}{MI-FGSM}   &   $\epsilon = 0.01 $ & 26.8 & 34.9  & \textbf{37.7}\\
 &  $\epsilon = 0.02 $ & 7.4 & 9.3 & \textbf{12.5} \\
\hline

\multirow{3}{*}{C\&W}   &  $c = 0.001 $ & 61.3 & 67.7 & \textbf{70.9} \\
 &  $c = 0.01 $ & 35.2 & 40.9 & \textbf{45.5} \\
 &  $c = 0.1 $ & 0.6 & 25.4 & \textbf{30.1} \\
\hline

\multirow{2}{*}{PGD}   & $\epsilon = 0.01 $ & 23.4 & 24.3 & \textbf{29.6} \\
&  $\epsilon = 0.02 $ & 6.0 & 7.8 & \textbf{10.2}\\
\hline \hline
\end{tabular}}

\end{center}
\end{table}

\begin{SCtable}[][h]
\centering
\caption{\small{\A{The convolutional neural network architecture used for adversarial training where the training set is augmented with adversarial and super-resolved images.}}}
\label{table:architecture}

\resizebox{0.675\width}{!}{
\begin{tabular}{c}
\hline \hline 

\cellcolor{black!10} \textbf{ResNet-110}\\
\hline \hline
 
 \multirow{3}{*}{
  \begin{tabular}{c}
  Conv(16, $3\times 3$) + BN\\
  ReLU($2\times 2$)
  \end{tabular}
  } \\ 
 \\
 \\
\hline

 \multirow{4}{*}{ {\Bigg[\Bigg[}
  \begin{tabular}{c}
  Conv(16$*k$, $1\times 1$) + BN\\
  Conv(16$*k$, $3\times 3$) + BN\\
  Conv(64$*k$, $1\times 1$) + BN\\
  \end{tabular}
 \Bigg] ${\times} 12$ \Bigg] $k {\in} \{1,2,4\}$}\\ 
\\
 \\
 \\
%\hline
%   \multirow{4}{*}{\Bigg[
%   \begin{tabular}{c}
%   Conv(32, $1\times 1$) + BN\\
%   Conv(32, $3\times 3$) + BN\\
%   Conv(128, $1\times 1$) + BN\\
%   \end{tabular}
%  \Bigg] $\times 12 $}\\

%  \\
%  \\
%  \\

% \hline

% \multirow{4}{*}{\Bigg[
%   \begin{tabular}{c}
%   Conv(64, $1\times 1$) + BN\\
%   Conv(64, $3\times 3$) + BN\\
%   Conv(256, $1\times 1$)+ BN\\
%   \end{tabular}
%  \Bigg] $\times 12 $}\\
% \\
%  \\
%  \\\
\hline
 \multirow{2}{*}{GAP}\\
 \\
\hline
 \multirow{2}{*}{FC(1024)}\\
\\
\hline
 \multirow{2}{*}{FC(10)} \\
\\
\hline
\end{tabular}}

\end{SCtable}

\subsection{Ablation Study}

\noindent \textbf{Super-resolution Methods:} Image super resolution recovers \textit{off-the-manifold} adversarial images from a low-resolution space and remaps them to the high-resolution space. This should hold true for different super-resolution techniques in the literature. In Table \ref{table:sr-comparison}, we evaluate the effectiveness of three image super-resolution techniques-  SR-ResNet, SR-GAN \cite{ledig2017photo} and EDSR \cite{lim2017enhanced}. Specifically, attacked images are super-resolved to $2\times$, without using any wavelet denoising. Experiments are performed on Inception v-3 classifier. The results in Table~\ref{table:sr-comparison} show a comparable performance across the evaluated super-resolution methods. These results demonstrate the effectiveness of super-resolution in recovering images. %It can be seen that each of these methods work equally good in defending against adversarial attacks. 

\begin{table*}[htp]
\caption{Individual contributions of Wavelet Denoising (WD) and Super Resolution (SR) towards the proposed defense scheme across three different classifiers.  Parameters: $\sigma_p=0.04$ and $S=2$. The proposed defense scheme works well across a range of classifiers.}
% \MH{Should we keep the first row? It shows degradation on clean images. Unless we show a comparison with other defenses, I'm of the opinion that we should remove 1st rows..}
% \A{I will run all the defenses on clean images today and report the comparison  }
\label{table:inception}
\centering
\begin{center}
\begin{tabular}{*{1}{l}| *{4}{c}| *{4}{c}| *{4}{c}r}
\hline
\rowcolor{black!10} \multicolumn{1}{l}{} & \multicolumn{4}{c}{\textbf{Inception v-3 model}} & \multicolumn{4}{c}{\textbf{ResNet-50 model}} & \multicolumn{4}{c}{\textbf{Inception ResNet v-2 model}}\\
\hline \hline
\textbf{Attack}              & \textbf{No Defense}  &\textbf{WD} &  \textbf{SR}  & \textbf{WD+SR} & \textbf{No Defense}  &\textbf{WD} &  \textbf{SR}  & \textbf{WD+SR} & \textbf{No Defense}  &\textbf{WD} &  \textbf{SR}  & \textbf{WD+SR}    \\
\hline
Clean            & 100 & 94.0&  96.2 &  97.0 & 100 & 92.7 & 93.2& 93.9   & 100 & 94.0& 97.2 & 98.2  \\
FGSM-2           & 31.7   & 57.3& 92.6   &  94.2 & 12.2  & 41.4&  85.4 & 86.1 & 59.4  & 70.5& 91.8  & 95.3   \\
FGSM-5           & 28.7   & 36.4& 85.7   &  87.9 & 7.0  & 12.7 & 74.0 & 77.2 & 55.0  & 57.5&   85.7 & 87.4 \\
FGSM-10          & 30.5  & 32.7 & 73.3  &  79.7 & 6.1  & 8.6&  60.5 & 64.9  & 53.6  &55.4&  79.4  & 82.3\\
I-FGSM           &  11.4 & 76.4 &  95.9 & 96.2  & 3.4 &  71.2 & 91.0 & 92.3 & 21.6  &  82.6 & 94.3&  95.8\\
DeepFool         & 0.4  &  74.9& 95.5   &  96.1 & 1.0  &71.8&   89.3 & 91.5  & 0.1  &79.1&   95.4 &  96.0 \\
C\&W             & 0.8   &76.3 & 95.6  &  96.0 & 0.1  & 79.0 &  92.0& 93.1 & 0.3  & 81.0 & 94.0 &  95.6 \\
MI-FGSM          & 1.7  &  77.0& 95.2   &  95.9  & 0.4  & 71.2 &    89.6 & 92.0 &  0.5 & 80.6 & 93.0 &  95.0  \\
DI$^2$FGSM       & 1.4  & 18.3 & 57.2   &  67.9  &  0.3 &  17.9&  49.8& 60.7 &  1.5 & 11.9 & 57.6 &  69.8  \\
MDI$^2$FGSM      & 0.6  & 5.8 & 27.1   &  31.7 &  0.2 &  9.4&  22.4 & 31.9 & 0.6  & 6.9 & 29.4 &  35.6  \\
\hline 
\end{tabular}
\end{center}
\end{table*}

Besides state-of-the-art image super-resolution methods, we further consider documenting the results on enhancing image resolution using interpolation-based techniques. For this, we perform experiments by resizing the images with Nearest Neighbor, Bi-linear and Bi-cubic interpolation techniques. In Table~\ref{table-nn-comparison}, we report the results achieved by three different strategies: upsample (by $2\times$), upsample + downsample and downsample + upsample. The results show that, although the performance of the simple interpolation based methods is inferior to more sophisticated state-of-the-art super-resolution techniques in Table~\ref{table:sr-comparison}, the simple interpolation based image resizing is surprisingly effective and achieves some degree of defense against adversarial attacks. \\ %Their performance is however inferior to more sophisticated and state-of-the-art super-resolution methods.%  Interpolation based methods though being computationally less intensive fail to perform well as well as deep learning based methods in defending against attacks.

\noindent \textbf{Effect of Wavelet Denoising:} Our proposed defense first deploys wavelet denoising, which aims to minimize the effect of adversarial perturbations, followed by image super-resolution to selectively introduce high-frequency components into an image (as seen in Fig. \ref{fig:filtering}) and recover \textit{off-the-manifold} attacked images. Here we investigate the individual impact of these two modules towards defending adversarial attacks. We perform extensive experiments on three classifiers: Inception v-3, ResNet-50 and InceptionResNet v-2. Table \ref{table:inception} shows the top-1 accuracy of each of the models for different adversarial attacks. The results show that, while wavelet denoising helps suppress added adversarial noise, the major performance boost is achieved with image super-resolution. The best performance is achieved when wavelet denoising is followed by super-resolution. These empirical evaluations demonstrate that image super-resolution with wavelet denoising is a robust model-agnostic defense technique for both iterative and non-iterative attacks. \\

\noindent \textbf{Hyper-parameters Selection:}
Unlike many existing defense schemes, which require computationally expensive model re-training and parameter optimization \cite{xie2017mitigating,kurakin2016adversarial,tramer2017ensemble,liao2017defense}, our proposed defense is training-free and does not require tuning a large set of hyper-parameters. Our proposed defense has two hyper-parameters: the scale of super-resolution ($S$) and the coefficient of BayesShrink ($\sigma_{\rho}$). We perform a linear search over the scaling factor $S$ for one single-step (FGSM-2) and one iterative (C\&W) attack on $500$ images, randomly selected from the ILSVRC validation set. These experiments are performed on Inception v-3 model. Table \ref{table:hyperparameters} shows the classifier performance across different super-resolution scaling factors. We select $S=2$, since it clearly shows significantly superior performance. Higher values of $S$ introduce significant high frequency components in the image, which degrade the performance. For $\sigma_{\rho}$, we follow \cite{prakash2018deflecting} and choose $\sigma_{\rho} \in \{0.03, 0.04, 0.05 \}$ as $\sigma_{\rho}=0.04$. \\

\noindent \response{ \textbf{Cross-Model Transferability Test:}
We further evaluate the \textit{black-box} settings which can be adapted closest to our approach as follows: \textit{firstly}, we generate adversarial examples using a source model, \textit{secondly}, we apply our image restoration techniques comprising of wavelet denoising and image super resolution and \textit{finally}, we test the images on a target model. The results in Table~\ref{table-cross-model-transferability} show the robustness of our method under these \textit{black-box} settings. We note a higher cross-model success rate for our defense under the above-mentioned settings.}\\

\begin{table}[t]
\caption{ \response{ \small{\textbf{Cross-Model Transferability Test:} PGD adversaries are first generated with $\epsilon = 8/255$, using the source network, followed by our image restoration scheme and then evaluated on target model. Note that the cross-model success rate of our defense is higher under \textit{black-box} settings. Numbers show robustness, higher the better.}} }
\label{table-cross-model-transferability}
\begin{center}
\resizebox{.45\textwidth}{!}{

\begin{tabular}{ c||c|c|c}
\hline \hline
\multirow{2}{*}{\diagbox{Source}{Target}} & \multirow{2}{*}{ResNet 50} & \multirow{2}{*}{Inception v-3} & \multirow{2}{*}{DenseNet 121}    \\
&  &  &  \\
\hline \hline
ResNet 50 & - & 84.7 & 78.0 \\
Inception v-3  & 78.0 & - & 75.8 \\ 
DenseNet 121  & 72.8 & 83.1 & - \\  
\hline
\end{tabular} }

\end{center}
\end{table}

%models \cite{xie2017mitigating,kurakin2016adversarial,tramer2017ensemble,liao2017defense} is use of fewer hyper-parameters which make it robust against all attack models. We in our experiments have two hyper-parameters: $S$, the scale of super-resolution and $\sigma_{\rho}$, coefficient of BayesShrink. We perform linear search over the scaling factor $S$ for one single-step (FGSM-2) and one iterative (C\&W) attack on $500$ images randomly chosen from the ImageNet validation-set. Table \ref{table:hyperparameters} shows the classifier performance across different super-resolution scaling factors. Following \cite{prakash2018deflecting} the hyper-parameter $\sigma_{\rho}$ is chosen from following set of values $\sigma_{\rho} \in \{0.03, 0.04, 0.05 \}$  and is empirically set at $0.04$.

\begin{table}[h]
\caption{ \small{Selection of super-resolution scaling factor. $S=2$ is selected due to its superior performance.}}
\label{table:hyperparameters}

\centering
\begin{center}
\begin{tabular}{lcccc}
\hline
\rowcolor{black!10} \textbf{Attack}    & \textbf{No Defense} & $S=2$  & $S=3$  & $S=4$ \\
\hline
Clean           & 100  & \textbf{97.2}  &  79.0  &  59.2 \\
FGSM            & 31.7 & \textbf{92.9}  &  76.2  &  58.8  \\
C\&W            & 0.3  & \textbf{95.8}  &  77.7  &  58.9 \\
\hline
\end{tabular}
\end{center}
\end{table}

\begin{SCfigure*}
\includegraphics[trim={0cm 5.7cm 8.2cm 0cm}, clip, width=0.835\textwidth]{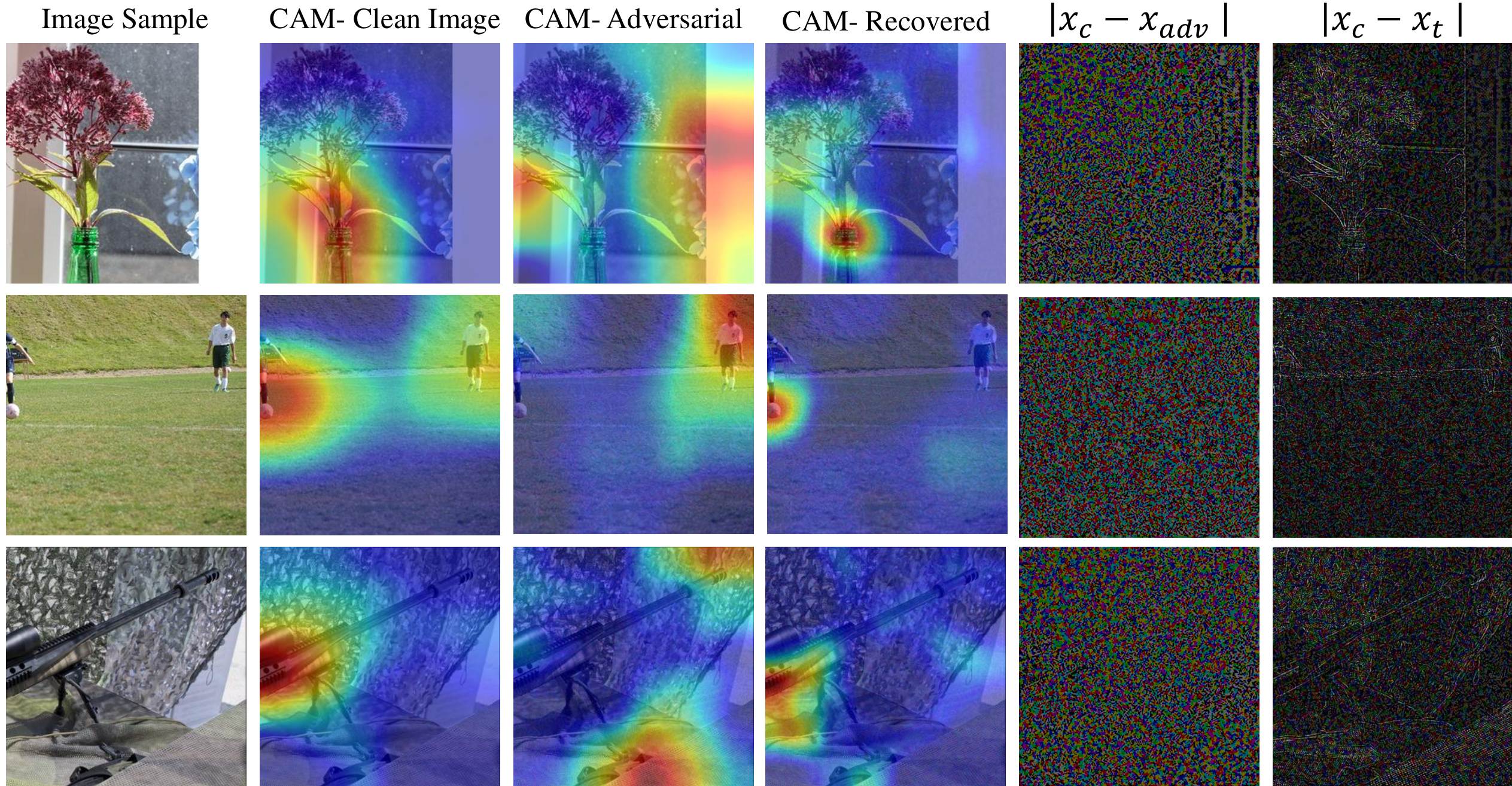}
\caption{\footnotesize{Visualization of Defense against Single-Step Attack (FGSM). First column shows three clean images. Subsequent three columns show the class activation maps for clean, FGSM ($\epsilon = 10$) attacked and recovered images. Second last column shows the perturbations (magnified $10x$) added to the clean image by FGSM and the last column shows the difference between clean image and defended image (magnified $10x$)}}\label{fig:fgsm}
\end{SCfigure*}

\begin{SCfigure*}
\includegraphics[trim={0cm 5.7cm 8.2cm 0cm}, clip, width=0.835\textwidth]{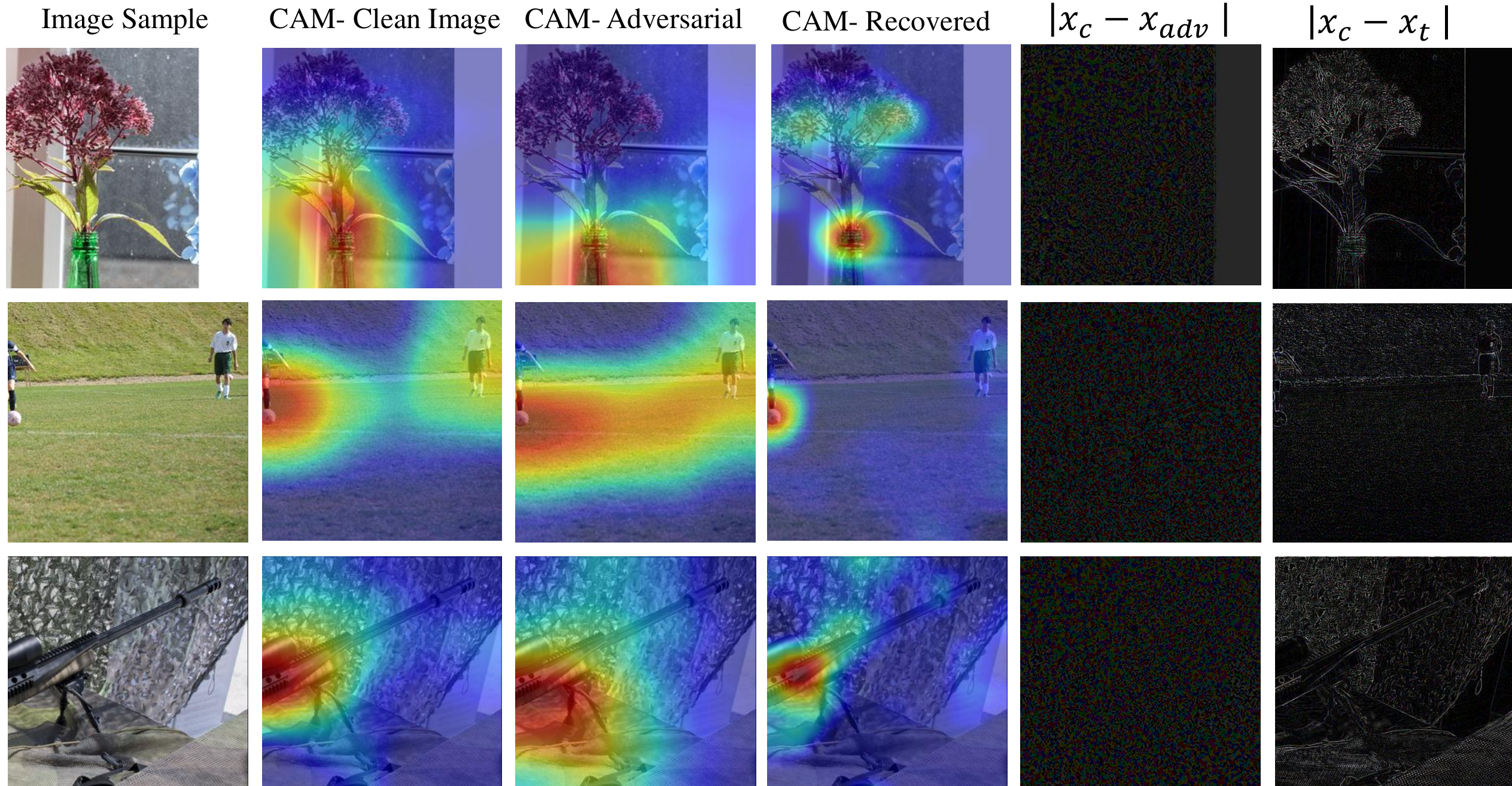}
\caption{\footnotesize{Visualization of Defense against Iterative Attack (CW). First column shows three clean images. Subsequent three columns show the class activation maps for clean, CW ($\ell_2$ norm) attacked and recovered images. Second last column shows the perturbations (magnified $40x$) added to the clean image by CW and the last column shows the difference between clean image and defended image (magnified $10x$)}}\label{fig:cw}
\end{SCfigure*}

\noindent  \textbf{CAMs Visualization:}
Class Activation Maps (CAMs)\cite{zhou2016learning} are weakly supervised localization techniques, which are helpful in interpreting the predictions of the CNN model by providing a visualization of discriminative regions in an image. CAMs are generated by replacing the last fully connected layer by a global average pooling (GAP) layer. A class weighted average of the outputs of the GAP results in a heat map which can localize the discriminative regions in the image responsible for the predicted class labels. Fig.~\ref{fig:fgsm} and ~\ref{fig:cw} show the CAMs for the top-1 prediction of Inception v-3 model for clean, attacked and recovered image samples. It can be observed that mapping an adversarial image to higher resolution destroys most of the noisy patterns, recovering CAMs similar to the clean images. Row 5 (Fig.~\ref{fig:fgsm} and ~\ref{fig:cw}) show the added perturbations to the clean image sample. Super-resolving an image selectively adds high-frequency components that eventually help in recovering model attention towards discriminative regions corresponding to the correct class labels (see Row 6, Fig.~\ref{fig:fgsm} and ~\ref{fig:cw}). 

\vspace{1em}

\begin{figure}[t]
  \centering
  \includegraphics[trim={0cm 1.05cm 9.1cm 0cm}, clip, width=0.49\textwidth]{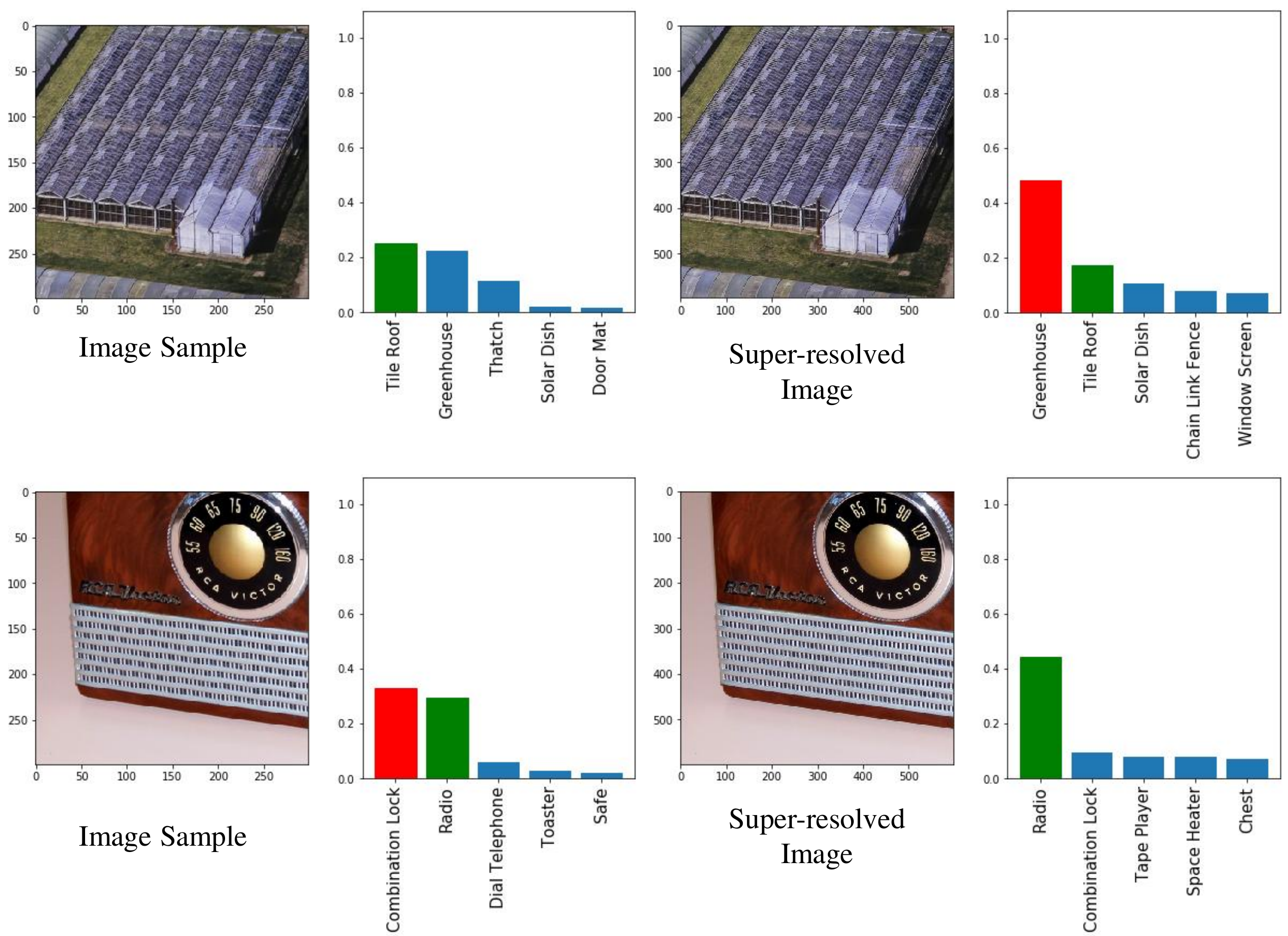}
  \caption{\small{\A{Effect of Image Super Resolution on clean image samples from the NIPS 2017 Dev Dataset. Predictions are made using a pre-trained Inception v-3 model. Green color refers to the correct class, while red and blue indicate incorrect classes. It can be seen that super resolving an image to a higher dimension at some instances depreciate the model's classification prediction (top), but can also recover the correct class for an otherwise misclassified clean image sample (bottom).}} }
  \label{fig:clean_vs_sr}
\end{figure}

\noindent  \textbf{Qualitative Analysis of SR:}
\A{In Fig.~\ref{fig:clean_vs_sr} we show two clean image samples where super-resolving the image to a higher dimension alters the classifier's predictions. In one case, super resolution causes the image to be misclassified. However, in the other, it recovers the actual class of the image which was otherwise incorrectly classified by a deep learning model. These cases predominantly arise in situations where the network's confidence for a single class is low, i.e. top-two predictions are roughly equal (see Fig.~\ref{fig:clean_vs_sr})}.

\begin{figure}[h]
  \centering
  \includegraphics[trim={0cm 8.79cm 18.0cm 0cm}, clip, width=0.4\textwidth]{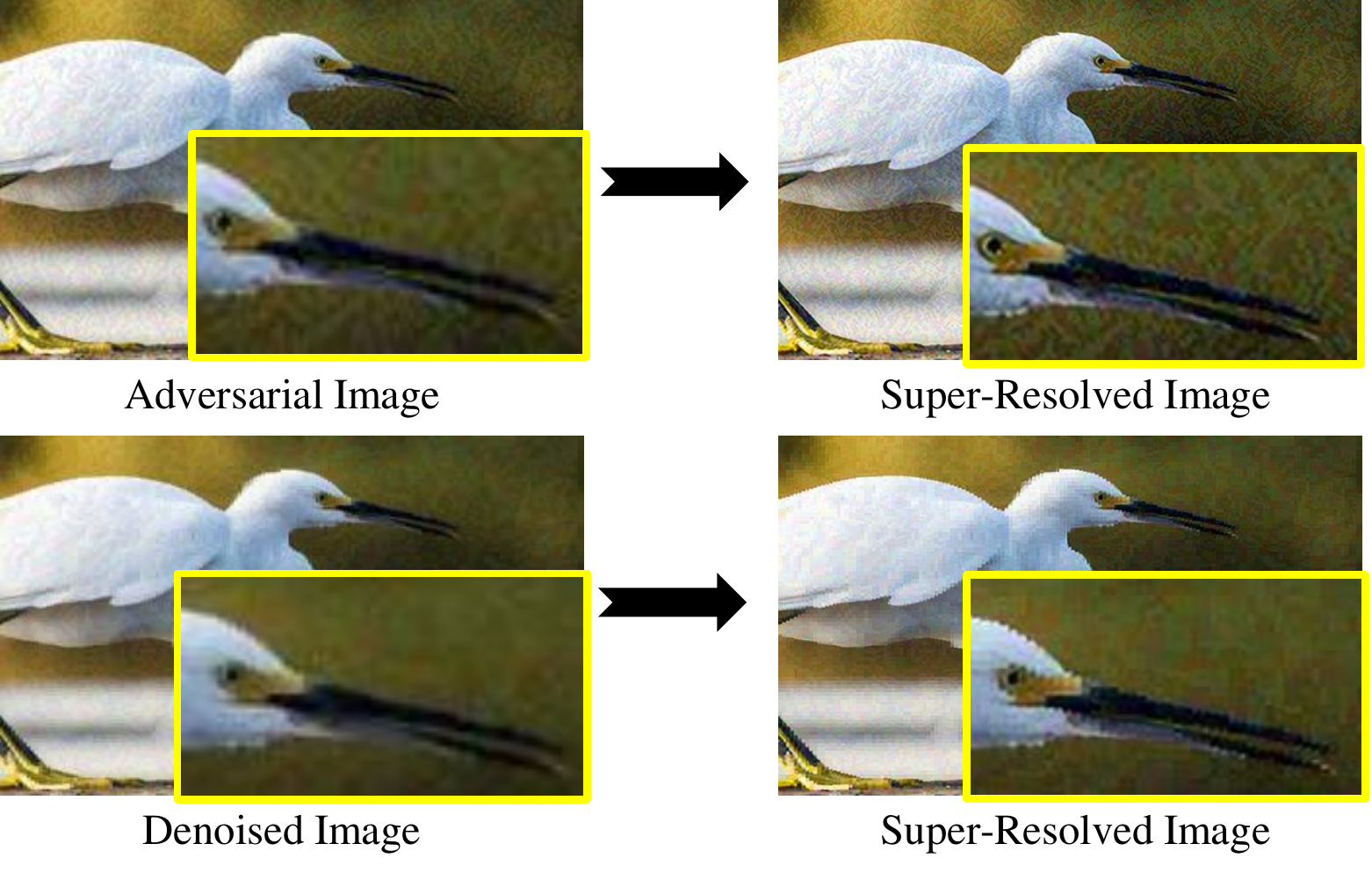}
  \caption{\small{\A{Qualitative effect of Image Super Resolution on adversarial image sample and its denoised counter-part.} }}
  \label{fig:denoised_sr}
\end{figure}

\A{In Fig.~\ref{fig:denoised_sr} we show the effect of individual components of the defense mechanism on a sample adversarial image. In an adversarial setting the perturbed images are generated with the objective of changing the model's prediction for an input without significantly changing the image (i.e., within a small bound $\epsilon$).  This means that the generated noise is not sampled from a predefined noise model, but is instead dependent on the loss surface and the input sample. Since SR models like EDSR may not deal with the noise, we first employ a denoiser in our proposed pipeline. As shown in Fig.~\ref{fig:denoised_sr}, the denoised inputs to the SR network perform reasonably well.}

% \A{The perturbed images are generated with the objective to change the model's prediction for an input without significantly changing the image (i.e., within a small bound $\epsilon$).  This means that the generated noise is not sampled from a predefined noise model, rather dependent on the loss surface and the input sample. Since, SR models like EDSR may not deal with the noise, we first employ a denoiser in our proposed pipeline. As shown in Fig.~\ref{fig:denoised_sr}, the denoised inputs to SR network perform reasonably well.
% }

\subsection{Identifying Obfuscated Gradients}

\A{Recently, Athalye \textit{et al.}\ \cite{Athalye2018ObfuscatedGG} were successful in breaking several defense mechanisms in the \textit{white-box} settings by identifying that they exhibit a false sense of security. They call this phenomenon \textit{gradient masking}. Below, we discuss how our defense mechanism does not cause gradient masking on the basis of characteristics defined in \cite{Athalye2018ObfuscatedGG, gilmer2018motivating}.}

\A{\noindent \textbf{Iterative attacks perform better than one-step attacks}: Our evaluations in Table~\ref{table-comparison} indicate that stronger iterative attacks (e.g. I-FGSM, MI-FGSM) are more successful at attacking the undefended models than single-step attacks (FGSM in our case).}

\A{\noindent \textbf{Robustness against \textit{gray-box} settings is higher than \textit{white-box} settings:} In \textit{white-box} settings, the adversary has complete knowledge of the model, so attacks should be more successful. In other words, if a defense does not suffer from obfuscated gradients, the robustness of the model against \textit{white-box} settings should be inferior to that in the \textit{gray-box} settings. In Table~\ref{table-white-box}, we show that the robustness under \textit{white-box} settings is lower than the robustness for \textit{gray-box} settings. This validates that the proposed defense follows the desired trend and does not obfuscate gradients.}

\begin{table}[h]

\caption{\small{\A{Performance of the proposed defense for NIPS 2017 Dev dataset when exposed to \textit{white-box} settings (in this case the adversary has complete knowledge of the denoising process and super-resolution model). Here $\epsilon$ is the perturbation size. }}}
% \vspace{-1em}
\label{table-white-box}
\begin{center}
\resizebox{.45\textwidth}{!}{
\begin{tabular}{c||c|c|c}
\hline 

\cellcolor{black!10} Attacks & \cellcolor{black!10} Params. &\cellcolor{black!10} \textit{White-box} &\cellcolor{black!10} \textit{Gray-box} \\

\hline \hline
\multirow{2}{*}{No Attack} & \multirow{2}{*}{-}   & \multirow{2}{*}{90.9} & \multirow{2}{*}{90.9}\\
 & & & \\
\hline
\multirow{3}{*}{FGSM}   & $\epsilon = 2/255 $ & 60.7 & 87.5  \\
 & $\epsilon = 5/255 $ & 49.5 & 79.9 \\
 & $\epsilon = 10/255 $ & 36.0 & 70.1 \\
\hline
\multirow{2}{*}{I-FGSM} & \multirow{2}{*}{$\epsilon = 16/255 $}   & \multirow{2}{*}{30.7} & \multirow{2}{*}{90.1}\\
 & & & \\
\hline
\multirow{2}{*}{MI-FGSM} & \multirow{2}{*}{$\epsilon = 16/255 $}   & \multirow{2}{*}{30.6} & \multirow{2}{*}{89.8}\\
 & & & \\
\hline\hline
\end{tabular}}

\end{center}
\end{table}

\begin{SCfigure}
\includegraphics[trim={0cm 0cm 0cm 0cm}, clip, width=0.315\textwidth]{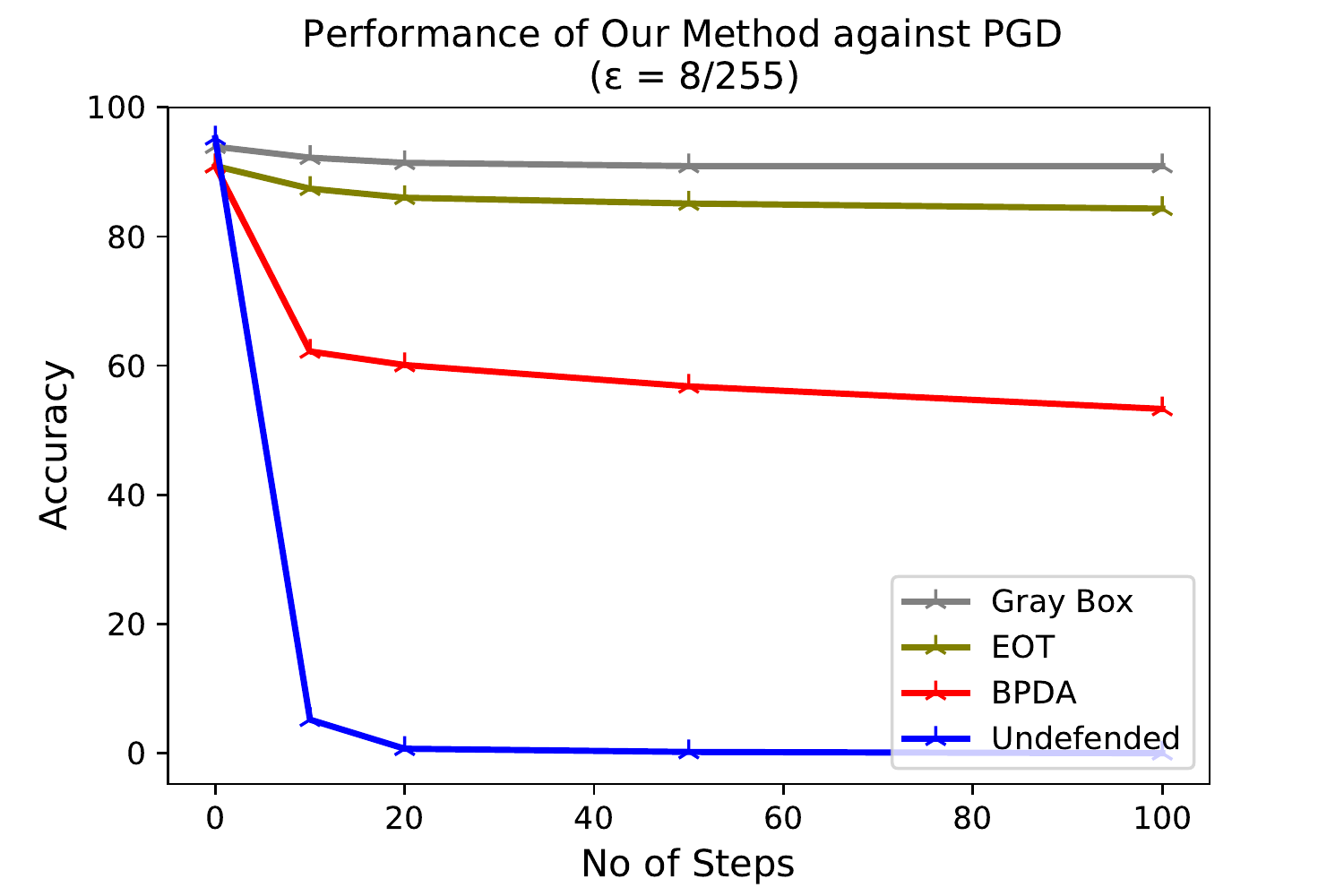}
\caption{\footnotesize{Performance of our defense model on adversarial images (NIPS 2017 dataset) generated using BPDA and EOT introduced by Athalye \textit{et al.} \cite{Athalye2018SynthesizingRA}. For an undefended model, the attack has a success rate of 100\%.}}\label{fig:graph-eot}
\end{SCfigure}

\A{\noindent  Since our defense scheme is based on a combination of transformations, wavelet-denoising and image super-resolution, we implement Backward Pass Differentiable Approximation (BPDA) to bypass the non-differentiable component of our defense. We also evaluate the robustness of our method against Expectation Over Transformation (EOT) \cite{Athalye2018ObfuscatedGG} attack. However, the attack methods fail to substantially break our defense, as shown in Fig.~\ref{fig:graph-eot}. With EOT \cite{Athalye2018ObfuscatedGG}, the accuracy drops by a mere 8.9\% for a strong attack (PGD) with a perturbation of $\epsilon = 8/255$.}

% \begin{figure}[h]
%   \centering
%   \includegraphics[trim={0cm 0cm 0cm 0cm}, clip, width=0.5\textwidth]{reply_images/Graph.pdf}
%   \caption{\small{\A{Performance of our defense model on adversarial images (NIPS 2017 dataset) generated using BPDA and EOT introduced by Athalye \textit{et al.} \cite{Athalye2018SynthesizingRA}. For an undefended model, the attack has a success rate of 100\%.} }}
%   \label{fig:graph-eot}
% \end{figure}

\section{Conclusion}
\label{sec:conclusion}
Adversarial perturbations can seriously compromise the security of deep learning based models. This can have wide repercussions since the recent success of deep learning has led to these models being deployed in a broad range of important applications, from health-care to surveillance. Thus, designing robust defense mechanisms that can counter adversarial attacks without degrading performance on unperturbed images is an absolute requisite. In this paper, we presented an image restoration scheme based on super-resolution, that maps \textit{off-the-manifold} adversarial samples back to the natural image manifold. We showed that the primary reason that super-resolution networks can negate the effect of adversarial noise is due to their addition of high-frequency information into the input image. Our proposed defense pipeline is agnostic to the underlying model and attack type, does not require any learning and operates equally well for \textit{black} and \textit{white-box} attacks. We demonstrated the effectiveness of the proposed defense approach compared to state-of-the-art defense schemes, where it outperformed competing models by a considerable margin. 

% \begin{figure*}[htp]
%   \centering
  
%   \includegraphics[trim={0cm 1.5cm 8.2cm 0cm}, clip, width=0.95\textwidth]{images/FGSM-CAM_v2.pdf}
%   \caption{\footnotesize{Visualization of Defense against Single-Step Attack (FGSM). First column shows four clean images. Subsequent three columns show the class activation maps for clean, FGSM ($\epsilon = 10$) attacked and recovered images. Second last column shows the perturbations (magnified $10x$) added to the clean image by FGSM and the last column shows the difference between clean image and defended image (magnified $10x$)}}
%   \label{fig:fgsm}
% \end{figure*}
%1.5

\bibliographystyle{IEEEtran}
\bibliography{root}

\end{document}